\documentclass{article} 
\usepackage{iclr2021_conference,times}

\usepackage{amsmath,amsfonts,bm}

\def\eqref#1{equation~\ref{#1}}

\def\1{\bm{1}}

\DeclareMathAlphabet{\mathsfit}{\encodingdefault}{\sfdefault}{m}{sl}
\SetMathAlphabet{\mathsfit}{bold}{\encodingdefault}{\sfdefault}{bx}{n}

\usepackage[utf8]{inputenc} 
\usepackage[T1]{fontenc}    
\usepackage{hyperref}       
\usepackage{url}            
\usepackage{booktabs}       
\usepackage{amsfonts}       
\usepackage{nicefrac}       
\usepackage{microtype}      
\usepackage[english]{babel}
\usepackage{blindtext}
\usepackage{siunitx}
\usepackage{graphicx}
\usepackage{svg}
\usepackage{multirow}
\usepackage{float}
\usepackage{placeins}
\usepackage{wrapfig}

\usepackage{xspace}
\usepackage[subtle,tracking=normal]{savetrees}

\newcommand{\beginsupplement}{
  \setcounter{table}{0}  
  \renewcommand{\thetable}{A\arabic{table}} 
  \setcounter{figure}{0} 
  \renewcommand{\thefigure}{A\arabic{figure}}
}

\title{Representation learning for improved interpretability and classification accuracy of clinical factors from EEG}

\author{%
Garrett Honke\thanks{Equal contribution} \textsuperscript{} \thanks{\texttt{\{ghonk,irinah,nthigpen,vmiskovic,katielink, sunnyd, pramodg\}@google.com}} \\
    X, the Moonshot Factory \\
    Mountain View, CA \\
    \And
Irina Higgins\footnotemark[1] \textsuperscript{} \footnotemark[2]\\
    DeepMind \\
    London, UK \\
    \And
Nina Thigpen\footnotemark[2] \\
    X, the Moonshot Factory \\
    Mountain View, CA, USA \\
    \And
Vladimir Miskovic\footnotemark[2] \\
    X, the Moonshot Factory \\
  Mountain View, CA, USA \\
  \And
Katie Link\footnotemark[2] \\
   X, the Moonshot Factory \\
  Mountain View, CA, USA \\
   \And
Sunny Duan\footnotemark[2] \\
   DeepMind \\
  Mountain View, CA, USA \\
    \And   
Pramod Gupta\footnotemark[2] \\
   X, the Moonshot Factory \\
  Mountain View, CA, USA \\
    \And
Julia Klawohn\thanks{JK is now at Humboldt-Universit\"{a}t zu Berlin, Berlin, Germany; \texttt{julia.klawohn@hu-berlin.de}} \\
   Florida State University \\
  Tallahassee, FL, USA \\
    \And
Greg Hajcak\thanks{\texttt{greg.hajcak@med.fsu.edu}} \\
   Florida State University \\
  Tallahassee, FL, USA \\
}

\arxiv
\begin{document}

\maketitle
\begin{abstract}
Despite extensive standardization, diagnostic interviews for mental health disorders encompass substantial subjective judgment. Previous studies have demonstrated that EEG-based neural measures can function as reliable objective correlates of depression, or even predictors of depression and its course. However, their clinical utility has not been fully realized because of 1) the lack of automated ways to deal with the inherent noise associated with EEG data at scale, and 2) the lack of knowledge of which aspects of the EEG signal may be markers of a clinical disorder. Here we adapt an unsupervised pipeline from the recent deep representation learning literature to address these problems by 1) learning a disentangled representation using $\beta$-VAE to denoise the signal, and 2) extracting interpretable features associated with a sparse set of clinical labels using a Symbol--Concept Association Network (SCAN). We demonstrate that our method is able to outperform the canonical hand-engineered baseline classification method on a number of factors, including participant age and depression diagnosis. Furthermore, our method recovers a representation that can be used to automatically extract denoised Event Related Potentials (ERPs) from novel, single EEG trajectories, and supports fast supervised re-mapping to various clinical labels, allowing clinicians to re-use a single EEG representation regardless of updates to the standardized diagnostic system. Finally, single factors of the learned disentangled representations often correspond to meaningful markers of clinical factors, as automatically detected by SCAN, allowing for human interpretability and post-hoc expert analysis of the recommendations made by the model.
\end{abstract}

\section{Introduction}
\label{sec_intro}
Mental health disorders make up one of the main causes of the overall disease burden worldwide \citep{vos2013global}, with depression (e.g., Major Depressive Disorder, MDD) believed to be the second leading cause of disability \citep{lozano2013global,whiteford2013global}, and around 17\% of the population experiencing its symptoms at some point throughout their lifetime \citep{McManus2016mental,McManus2009adult,kessler1993sex,lim2018prevalence}. At the same time diagnosing mental health disorders has many well-identified limitations \citep{insel2010research}. Despite the existence of diagnostic manuals like Structured Clinical Interview for Diagnostic and Statistical Manual of Mental Disorders (SCID) \citep{dsm5}, diagnostic consistency between expert psychiatrists and psychologists with decades of professional training can be low, resulting in different diagnoses in upwards of 30\% of the cases (Cohen’s Kappa = 0.66) \citep{lobbestael2011inter}. Even if higher inter-rater reliability was achieved, many psychological disorders do not have a fixed symptom profile, with depression alone having many hundreds of possible symptom combinations \citep{fried2015depression}. This means that any two people with the same SCID diagnosis can exhibit entirely different symptom expressions. This is a core challenge for developing an objective, symptom-driven diagnostic tool in this domain.

\begin{figure}[!t]
\vskip -0.3in
  \centering
  \includegraphics[width=1.\textwidth]{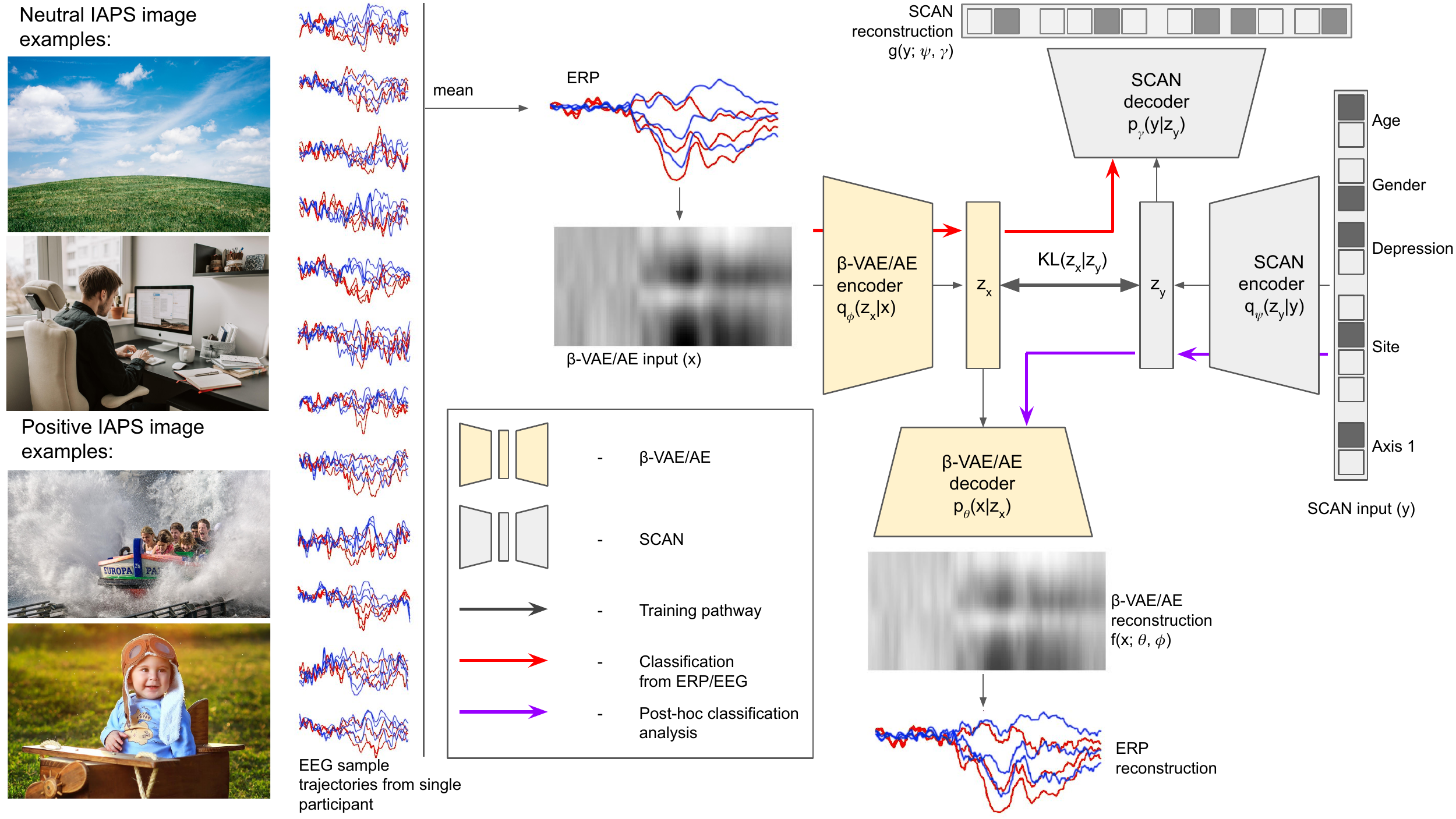}
  \vskip -0.1in
\caption{Pipeline schematic. Participants are presented with images from the International Affective Picture System (IAPS). EEG trajectories recorded from the same participant over multiple trials are averaged to create ERPs. Each EEG sample consists of 256 time samples (-248-772~ms, where 0 is stimulus onset time) \color{black} of stimulus-locked activity recorded from three sites as participants view either neutral (red)\color{black} or positive (blue) IAPS images. ERP responses to positive and negative images are concatenated and normalised between [0, 1] across all channels simultaneously\color{black}. The resulting ``images'' are used to train $\beta$-VAE or AE. A well disentangled pre-trained $\beta$-VAE model is used to train SCAN. Each SCAN training example consists of a 5-hot binary classification label $\textbf{y}$ presented to the SCAN encoder, and its corresponding EEG ``image'' $\textbf{x}$ presented to the $\beta$-VAE encoder.  $\beta$-VAE weights are fixed during SCAN training. To obtain classification, an EEG ``image'' is presented to the $\beta$-VAE encoder, then the inferred $\textbf{z}_x$ means are fed through the SCAN decoder, where a per-class softmax is applied to obtain the predicted label (red pathway). To analyse SCAN classification decisions, a 1-hot binary vector is fed into SCAN encoder. Samples from the inferred distribution $\textbf{z}_y$ are then fed through the $\beta$-VAE decoder to visualise the corresponding ERP reconstructions (purple pathway). }
  \label{fig_model}
  \vskip -0.2in
\end{figure}

Electroencephalography (EEG) is a measurement of post-synaptic electrical potentials that can be taken non-invasively at the scalp. EEG signals can function as important biomarkers of clinical disorders \citep{hajcak2019utility} but they are difficult to clean and interpret at scale. For example, components of the EEG signal can often significantly overlap or interfere with each other. Furthermore, nearby electronics, line noise, hardware quality, signal drift and other variations in the electrode--scalp connection can all distort the recorded EEG signal. Hence, the extraction of EEG data of sufficient quality is usually a laborious, semi-automated process executed by lab technicians with extensive training. A typical EEG analysis pipeline consists of collecting EEG recordings evoked from a large number of stimulus presentations (trials) in order to have sufficient data to average out the noise. Independent Components Analysis (ICA) is often used to visually identify and remove the component that corresponds to eye blinks \citep{delorme2004eeglab, makeig2004mining, jung2000removing}. This can be followed by a trial rejection stage where anomalous trials are identified and removed from the EEG data scroll, sometimes also through visual examination. The cleaned up EEG recordings from a large number of trials are then averaged to produce an Event Related Potential (ERP) \citep{luck2012event}. This allows a clinician to extract specific ERP components relevant to the clinical factor of interest, average out the event-locked activity within them,  and then either perform a statistical group comparison, or---in the case of the diagnosis classification goal---apply an off-the-shelf classifier, like Logistic Regression (LR) to obtain the final diagnostic results. Some more advanced classification approaches might include Support Vector Machines (SVM), Linear Discriminant Analysis (LDA), or Random Forest (RF) \citep{parvar2015detection, guler2007multiclass, subasi2010eeg, tomioka2007logistic, bashivan2016learning}. 

To summarise, EEG recordings are noisy measures of electric activity from across the brain. There is evidence that these signals are useful as markers of depression, but we lack understanding of what aspects of depression they index. Furthermore, the field of clinical psychopathology still lacks consensus on the etiopathogenesis of mental health disorders, which means that there is no such thing as the ``ground truth'' diagnostic labels. Hence, while EEG is routinely used to diagnose conditions like epilepsy \citep{Smith2005eeg}, memory \citep{stam1994non} or sleep disorders \citep{karakis2012nonlinear}, its promise for being a reliable diagnostic tool for clinical conditions like depression has not been fully realised so far. In order to make EEG a viable diagnostic tool for a broader set of clinical conditions it is important to develop an automated pipeline for extracting the relevant interpretable biomarker correlates from the (preferably individual trial) EEG data in a robust manner. Furthermore, this process should not depend fully on diagnostic labels which are often subjective and at best represent highly heterogeneous classes. 

Recent advances in deep learning have prompted research into end-to-end classification of EEG signal using convolutional and/or recurrent neural networks \citep{bashivan2016learning, mirowski2009classification, cecotti2011convolutional, guler2005recurrent, wang2018convolutional, farahat2019convolutional, solon2019decoding, cecotti2014single}, holding the promise of automated extraction of the relevant biomarker correlates. However, deep classifiers operate best in the big data regime with clean, well-balanced ground truth classification targets. In contrast, even the largest of EEG datasets typically contain only a few hundred datapoints, and the classification labels are subjective, noisy and unbalanced, with the majority of the data coming from healthy control participants. Hence, in order to utilise the benefits of deep learning but avoid the pitfalls of over-reliance on the classification labels, we propose a two-step pipeline consisting of unsupervised representation learning, followed by supervised mapping of the pre-trained representation to the latest version of the available diagnostic labels. The hope is that the unsupervised learning step would denoise the input signal and extract the broad statistical regularities hidden in it, resulting in a representation that can continue to be useful even if the label taxonomy evolves. 

Recently great progress has been made in the field of deep unsupervised representation learning \citep{roy2019deep, devlin2018bert, brown2020language, chen2020simple, grill2020bootstrap, chen2020generative, higgins2017beta, burgess2019monet}\color{black}. Disentangled representation learning is a branch of deep unsupervised learning that produces interpretable factorised low-dimensional representations of the training data \citep{bengio2013better, higgins2017beta}. Given the requirement for model interpretability in our use-case, we use Beta Variational Autoencoders ($\beta$-VAE) \citep{higgins2017beta}---one of the state of the art unsupervised disentangled representation learning methods---to discover low-dimensional disentangled representations of the EEG data. We then train the recently proposed Symbol--Concept Association Network (SCAN) \citep{higgins2017scan} to map the available classification labels to the representations learnt by $\beta$-VAE (see Fig.~\ref{fig_model}). We demonstrate that our proposed pipeline results in better classification accuracy than the heavily hand-engineered baseline typical for the field. This holds true when predicting a number of factors, including age, gender, and depression diagnosis. Furthermore, SCAN is able to produce highly interpretable classification recommendations, whereby its decisions on different clinical factors are grounded in a small number (often single) latent dimensions of the $\beta$-VAE, allowing the clinicians an opportunity to interpret the recommendations produced by SCAN, and visualise what aspects of the EEG signal are associated with the classification decision post-hoc. This opens up the opportunity to use our proposed pipeline as a tool for discovering new EEG biomarkers. We validate this by ``re-discovering'' a known biomarker for depression. Finally, we demonstrate that once a $\beta$-VAE is pre-trained on ERP signals, it can often produce ERP-like reconstructions even when presented with single noisy EEG trajectories. Furthermore, the representations inferred from single EEG trials produce good classification results, still outperforming the canonical baseline method. This suggests that once a good disentangled representation is learnt, the model can be used online as new EEG data is being recorded, thus lowering the burden of keeping potentially vulnerable participants in the lab for extended recording sessions.

\section{Methods}

\subsection{Data}
\paragraph{Anhedonia.}
This work targets one of the two cardinal symptoms of depression---anhedonia. Anhedonia is the lack of pleasure and/or interest in previously pleasurable stimuli and activities \citep{dsm5}. One established approach for objectively quantifying this symptom is the use of EEG to measure neural responses elicited by emotionally salient visual stimuli. Research in this domain has uncovered a stereotyped neural activation pattern in healthy control participants, where emotionally-salient stimuli evoke a Late Positive Potential (LPP) in ERPs---the averaged timeseries of stimulus time-locked EEG recordings. This pattern has been identified as a potential biomarker for depression because (on average) this positive deflection in amplitude is attenuated or absent in individuals who exhibit symptoms of anhedonia
\citep{brush2018using, foti2010reduced, klawohn2020reduced, macnamara2016diagnostic, weinberg2016depression, weinberg2017blunted}.

\paragraph{Participants.} 
The data were collected as part of a mental health study across multiple laboratory sites. The multi-site aspect of the study meant that more data could be pooled together, however, it also meant that the data was noisier. Participants ($N=758,\ \textup{age}_{\bar{X}}=16.7,\ \textup{age}_{\textup{range}}=[11.0,\ 59.8]$, 398 female) were samples of healthy controls ($n_{\textup{HC}}=485$) and people diagnosed with anxiety and depression (among other mental illnesses) (see Sec.~\ref{app_sec_participants} and Tbl.~\ref{tbl_participant_demos} for further breakdown).

\paragraph{Stimuli and Experimental Design.} 
Participants were shown a series of 80 images from the International Affective Picture System (IAPS) \citep{lang2008international} presented in random order up to 40 times each. The images varied in valence: either positive (affiliative scenes or cute animals designed to elicit the LPP ERP component), or neutral (objects or scenes with people). Each image trial consisted of a white fixation cross presented for a random duration between 1000-2000~ms (square window) followed by a black and white image presented for 1000~ms.

\paragraph{EEG Preprocessing.} 
While participants completed the picture viewing task, EEG was continuously recorded. Each picture trial was then segmented to contain a 200~ms pre-stimulus baseline and a 1200~ms post-stimulus interval. The raw EEG signal was digitized, bandpass filtered and cleared of the eye movement artifacts and anomalous trials as described in Sec.~\ref{app_eeg_preprocessing}.

\paragraph{Classification labels.} 
The following classification labels were used in this study: age (adult or child), gender (male or female), study site, and the presence or absence of two clinical conditions: depression diagnosis and a broader Axis 1 disorder diagnosis. All classification labels were binary, apart from study site, which contained four possible values corresponding to four different sites where the data were collected. Participants 18 years of age and older were classified as adults. Gender was classified based on self-reported values. Positive depression labels ($n=110$) include all participants that were diagnosed with Major Depressive Disorder (MDD), Persistent Depressive Disorder (PDD), and depressive disorder NOS (not-otherwise-specified) by expert clinicians through a clinical interview (e.g., SCID for adults, KSADS for children). Axis 1 is a broad category of disorders (now discontinued in DSM-V) that excludes intellectual disabilities and personality disorder \citep{dsm5}. Positive Axis 1 labels ($n=273$)  encompassed all participants with positive depression labels plus individuals diagnosed with Cyclothemia, Dysthemia, anxiety disorders, mood disorders (e.g., Bipolar Disorder), panic disorders, and substance and eating disorders (all of which are sparse).\footnote{Disorders included in the Axis 1 label that occur in the data include Major Depressive Disorder, Persistent Depressive Disorder, Depression NOS, Cyclothemia, Dysthemia, Bipolar I, Bipolar II, Bipolar NOS, Mania, Hypomania, Agoraphobia, Social Phobia, Separation Anxiety, Generalized Anxiety Disorder, Panic Disorder, Panic Disorder with Agoraphobia, Anorexia, Bulimia, Eating disorder NOS, Alcohol Abuse, Alcohol Dependence, and Substance Abuse and Substance Dependence disorders.}


\subsection{Representation learning}
\paragraph{Canonical LPP analysis baseline}
The canonical approach for extracting the Late Positive Potential (LPP) effect serves as the baseline for this work. The LPP effect is calculated as the average amplitude difference between ERP waveforms (i.e., averaged stimulus time-locked EEG segments) evoked from emotionally-salient and neutral stimuli. Before the delta was calculated, each ERP was normalised with respect to the baseline average activity within the 100~ms window preceding stimulus onset. Finally, the normalised delta signal within the 300–700~ms window after stimulus onset was averaged to provide the canonical baseline LPP representation.

\paragraph{Autoencoder}
An AutoEncoder (AE) is a deep neural network approach for non-linear dimensionality reduction \citep{hinton2006reducing, baldi2012autoencoders}. A typical architecture consists of an encoder network, which projects the input high-dimensional data $\mathbf{x}$ into a low-dimensional representation $\mathbf{z}$, and a decoder network that is the inverse of the encoder, projecting the representation $\mathbf{z}$ back into the reconstruction of the original high-dimensional data $f(\mathbf{x};\ \phi, \theta)$, where $\phi$ and $\theta$ are the parameters of the encoder and decoder respectively (see Fig.~\ref{fig_model} for model schematic). AE is trained through backpropagation \citep{rumelhart1986learning} using the reconstruction objective:

\begin{equation*}
    \mathcal{L}_{AE} = \mathbb{E}_{p(\mathbf{x})} \ || f(\mathbf{x};\ \phi, \theta) - \mathbf{x} ||^2
\end{equation*}

The input to the AE in our case is a 256x6 ``image'' of the EEG signal (see Fig.~\ref{fig_model}), where 256 corresponds to the 1024~ms of the recorded EEG trajectory sampled at 250~Hz and pre-processed as described in Sec.~\ref{app_eeg_preprocessing}, and 6 corresponds to three electrodes per each of the two image valence conditions (i.e., stimulus classes): neutral and positive. 
These input images were further normalised to the [0, 1] range  across all channels before being presented to the AE.\color{black} 

 The AE was parametrised to have two convolutional encoding layers with 32 filters each of size 6, with strides of step size 2 along the time axis, followed by a single fully connected layer of size 128, projecting into a 10-dimensional representation $\mathbf{z}$. The decoder was the inverse of the encoder. Overall the model consisted of around 106,752 parameters. The model had ReLU activations throughout, and was optimised using Adam optimizer with learning rate of 1e-4 over 1~mln iterations of batch size 16.\color{black}

\paragraph{$\beta$-Variational Autoencoder}
A $\beta$-Variational Autoencoder ($\beta$-VAE) \citep{higgins2017beta} is a generative model that aims to learn a disentangled latent representation $\mathbf{z}$ of input data $\mathbf{x}$ by augmenting the Variational Autoencoder (VAE) framework \citep{rezende2014stochastic,kingma2013auto} (see Sec.\ref{app_vae} for more details) with an additional $\beta$ hyperparameter. Intuitively, a disentangled representation is a factorised latent distribution where each factor corresponds to an interpretable transformation of the training data (e.g. in a disentangled representation of a visual scene, individual factors may represent a change in lighting or object position). A neural network implementation of a $\beta$-VAE consists of an inference network (equivalent to the AE encoder), that takes inputs $\mathbf{x}$ and parameterises the (disentangled) posterior distribution $q(\mathbf{z}|\mathbf{x})$, and a generative network (equivalent to the AE decoder) that takes a sample from the inferred posterior distribution $\hat{\mathbf{z}} \sim \mathcal{N}(\mu(\mathbf{z}|\mathbf{x}), \sigma(\mathbf{z}|\mathbf{x}))$ and attempts to reconstruct the original image (see Fig.~\ref{fig_model}). The model is trained through a two-part loss objective:

\begin{equation*}
     \mathcal{L}_{\ensuremath{\beta}-VAE} = \mathbb{E}_{p(\mathbf{x})} [\ \mathbb{E}_{q_{\phi}(\mathbf{z}|\mathbf{x})} [\log\ p_{\theta}(\mathbf{x} | \mathbf{z})] - \beta KL(q_{\phi}(\mathbf{z}|\mathbf{x})\ ||\ p(\mathbf{z}))\ ]
\end{equation*}

where $p(\mathbf{x})$ is the probability of the input data, $q(\mathbf{z}|\mathbf{x})$ is the learnt posterior over the latent units given the data, $p(\mathbf{z})$ is the unit Gaussian prior with a diagonal covariance matrix $\mathcal{N}(\mathbf{0}, \mathbb{I})$, $\phi$ and $\theta$ are the parameters of the inference (encoder) and generative (decoder) networks respectively, and $\beta$ is a hyperparameter that controls the degree of disentangling achieved by the model during training. Intuitively, the objective consists of the reconstruction term (which aims to increase the log-likelihood of the observations) and a compression term (which aims to reduce the Kullback-Leibler (KL) divergence between the inferred posterior and the prior). Typically a $\beta>1$ is necessary to achieve good disentangling, however the exact value differs for different datasets. In order to find a good value of $\beta$ to disentangle our EEG dataset, we perform a hyperparameter search, training ten models with different random initialisations for each of the ten values of $\beta \in [0.075,\ 2.0]$ sampled uniformly. Well disentangled $\beta$-VAE models were selected using the Unsupervised Disentanglement Ranking (UDR) score \citep{duan2019unsupervised} described in Sec.~\ref{app_udr} (see Fig.~\ref{fig_udr_scores} for a visualisation of the resulting UDR scores). All $\beta$-VAE models had the same architecture as the AE\footnote{Note that the AE can be seen as a special case of the $\beta$-VAE with $\beta=0$.},  and were trained in the same manner and on the same data that was pre-processed in the same way to lie in the [0, 1] range \color{black}.

\begin{figure}[t!]
\vskip -0.3in
  \centering
  \includegraphics[width=1.\textwidth]{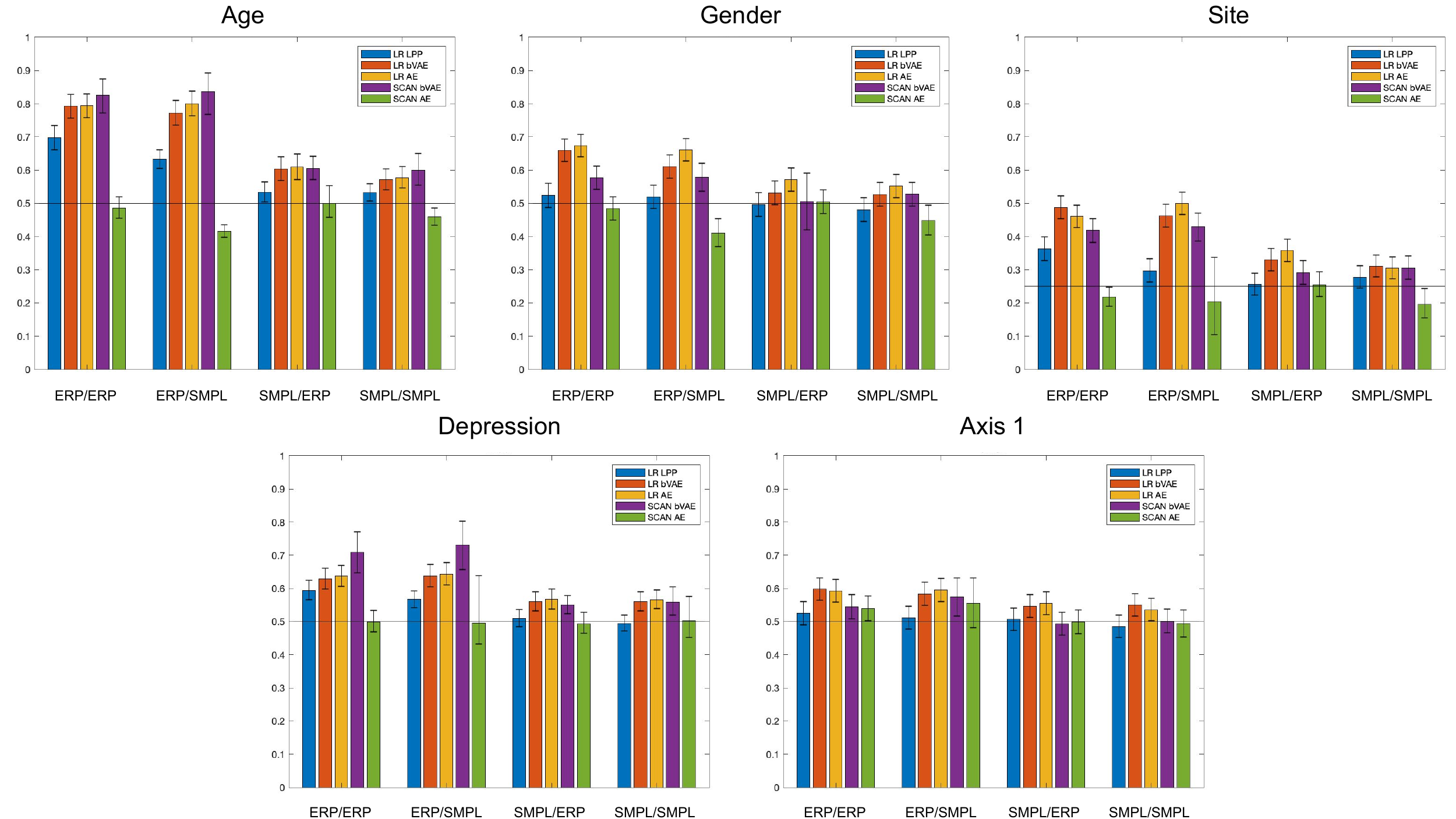}  
  \vskip -0.1in
  \caption{ Posterior distribution of balanced classification accuracy (mean, error bars - 95\% confidence interval) \citep{carrillo2014probabilistic,brodersen2010balanced}. ERP - averaged trajectories, SMPL - single sampled EEG trials. LPP - canonical late positive potential baseline representation. LR - L2 regularised logistic regression (see Tbl.~\ref{tbl_classification_results_baselines} for other baseline classifiers). Models trained on single EEG trajectories (SMPL) are tested on different single EEG trajectories in the SMPL/SMPL train/test case. Black horizontal line - chance accuracy.\color{black} }
  \label{fig_classification_results}
  \vskip -0.1in
\end{figure}

\subsection{Classification}
\paragraph{Baseline classifiers}
To evaluate the quality of a representation in terms of how useful it is for classifying different clinical factors, we applied a range of baseline classification algorithms: Support Vector Machine (SVM), Random Forest (RF), Logistic Regression (LR) and Linear Discriminant Analysis (LDA) (see Sec.~\ref{app_classifiers} for details). For all classification results we report the posterior distribution of balanced accuracy \citep{carrillo2014probabilistic,brodersen2010balanced}. Balanced accuracy was chosen, because it correctly matches chance accuracy even for unbalanced datasets.

\paragraph{Symbol--Concept Association Network}
The baseline classifiers described above produced uninterpretable decisions. This is undesirable if we were to have a chance at discovering new clinical biomarkers in the EEG data. To address this interpretability challenge, we leverage a recent model proposed for visual concept learning in the machine learning literature---the Symbol--Concept Association Network (SCAN) \citep{higgins2017scan}. While SCAN was not originally developed with the classification goal in mind, it has desirable properties to utilise for the current application. In particular, it is able to automatically discover sparse associative relationships between discrete symbols (5-hot classification labels in our case, see Fig.~\ref{fig_model}) and continuous probability distributions of the disentangled posterior representation discovered by a trained $\beta$-VAE model. Furthermore, the associative nature of the grounding used to train SCAN allows it to deal with noisy data gracefully, and to learn successfully from a small number of positive examples and from highly unbalanced datasets.  
SCAN is in effect another VAE model. In our case it takes 5-hot classification labels $\mathbf{y}$ as input, and aims to reconstruct them from the inferred posterior $q(\mathbf{z}|\mathbf{y})$ (see Fig.~\ref{fig_model} 
for more details). To train SCAN the original VAE objective is augmented with an additional KL term that aims to ground the SCAN posterior in the posterior of a pre-trained $\beta$-VAE model:

\begin{equation*}
     \mathcal{L}_{SCAN} = \mathbb{E}_{p(\mathbf{y})} [\ \mathbb{E}_{q_{\psi}(\mathbf{z}_y|\mathbf{y})} [\log\ p_{\gamma}(\mathbf{y} | \mathbf{z}_y)] - KL(q_{\psi}(\mathbf{z}_y|\mathbf{y})\ ||\ p(\mathbf{z}_y))\ ] - KL( q_{\phi}(\mathbf{z}_x|\mathbf{x})\ ||\ q_{\psi}(\mathbf{z}_y|\mathbf{y}) )\ ]
\end{equation*}

where $\psi$ and $\gamma$ are the parameters of the SCAN encoder and decoder respectively, $q_{\psi}(\mathbf{z}_y|\mathbf{y})$ is the posterior of SCAN inferred from a 5-hot label $\mathbf{y}$, and $q_{\phi}(\mathbf{z}_x|\mathbf{x})$ is the posterior of the pre-trained $\beta$-VAE model inferred from an EEG ``image'' corresponding to the label presented to SCAN. Note that the $\beta$-VAE weights are not updated during SCAN training. The extra grounding term $KL( q_{\phi}(\mathbf{z}_x|\mathbf{x})\ ||\ q_{\psi}(\mathbf{z}_y|\mathbf{y}) )$ allows SCAN to discover an associative relationship between a subset of disentangled factors and each 1-hot dimension of the given label. 

 We paramterised SCAN encoder and decoder as MLPs with two hidden layers of size 128, ReLU non-linearities and a 10-dimensional posterior to match the dimensionality of the $\beta$-VAE representation. The model resulted in around 22,016 parameters. Like the other models, SCAN was trained with batch size 16, over 1~mln iterations and with Adam optimizer with learning rate of 1e-4.\color{black}

\section{Results}

\begin{figure}[t!]
\vskip -0.3in
  \centering
  \includegraphics[width=1.\textwidth]{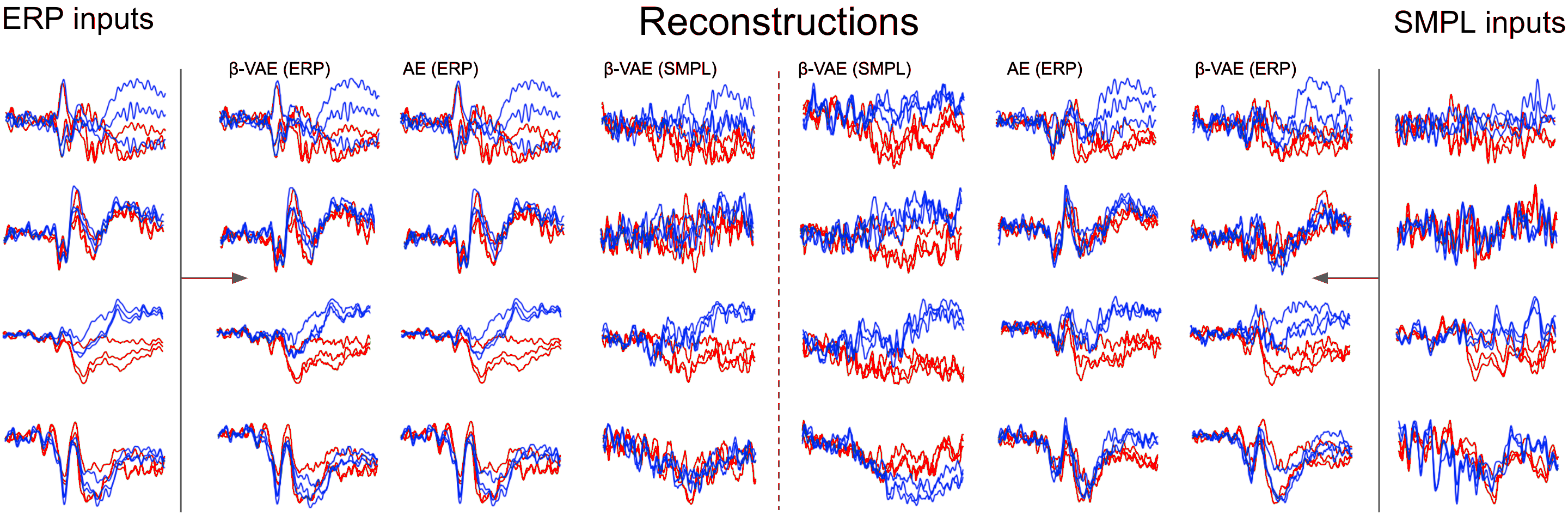}
  \vskip -0.1in
  \caption{Reconstructions (middle six columns) by two disentangled $\beta$-VAEs pre-trained on either ERPs or single EEG sample trajectories (SMPL), as well as an AE pre-trained on ERPs. Each model reconstructs either single EEG trajectories (SMPL inputs) or the corresponding ERPs (ERP inputs). Models pre-trained on ERPs are able to reconstruct ERP-like trajectories even from single EEG samples, as demonstrated by the closer similarity of the $\beta$-VAEs (ERP) and AE (ERP) reconstructions to ERP inputs compared to SMPL inputs, regardless of whether ERPs or single EEG samples are reconstructed. }
  \label{fig_sample_reconstructions}
  \vskip -0.1in
\end{figure}

\begin{figure}[t!]
\vskip -0.3in
  \centering
  \includegraphics[width=1.\textwidth]{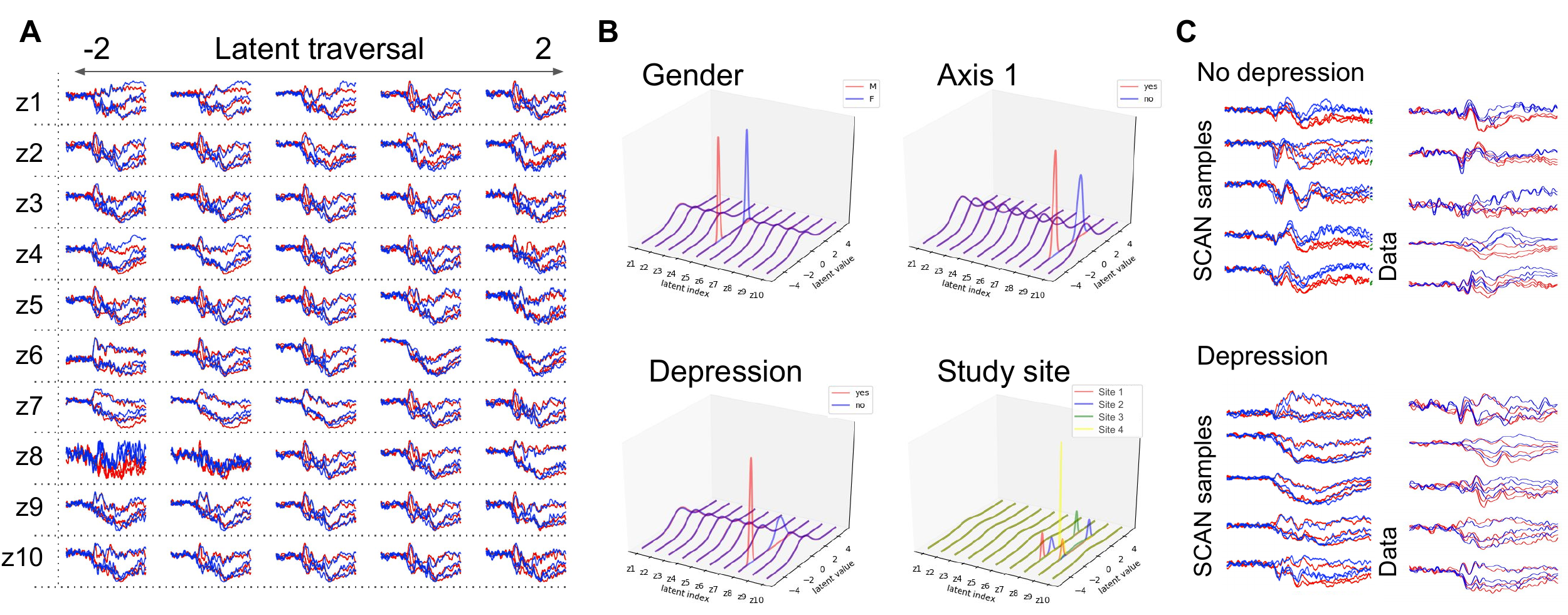}
  \vskip -0.1in
  \caption{\textbf{A}: Latent traversals of a pre-trained disentangled $\beta$-VAE. Each row reconstructs an ERP trajectory as the value of each latent dimension is traversed between [-2, 2] while keeping the values of all other latents fixed. \textbf{B}: Inferred distributions for each of the 10 latent dimensions of a SCAN trained with the $\beta$-VAE shown in A. Each subplot shows the result of SCAN inference on a set of 1-hot symbols corresponding to all the different conditions of a single factor (e.g. M/F for Gender). Different factors have automatically become associated with different (often single) latent dimensions. For example, the presence (red) or absence (blue) of Axis 1 diagnosis is represented by the narrow inferred Gaussian distributions centered around different opposite values of latent $z_9$, while other latent dimensions are defaulted back to the unit Gaussian prior. \textbf{C}: five samples from each of the two conditions associated with the depression factor from the SCAN shown in B,   as well as corresponding ERP samples from the training dataset \color{black}. Red - neutral responses, blue - pleasant responses from each of the three EEG sensors. SCAN has automatically discovered that the presence of depression is associated with the lack of clear separation between responses in the two valence conditions---anhedonia. Its samples are qualitatively similar to the samples from the dataset. }
  \label{fig_traversals}
  \vskip -0.1in
\end{figure}

\begin{wrapfigure}{r}{0.5\textwidth}
  \includegraphics[width=0.48\textwidth]{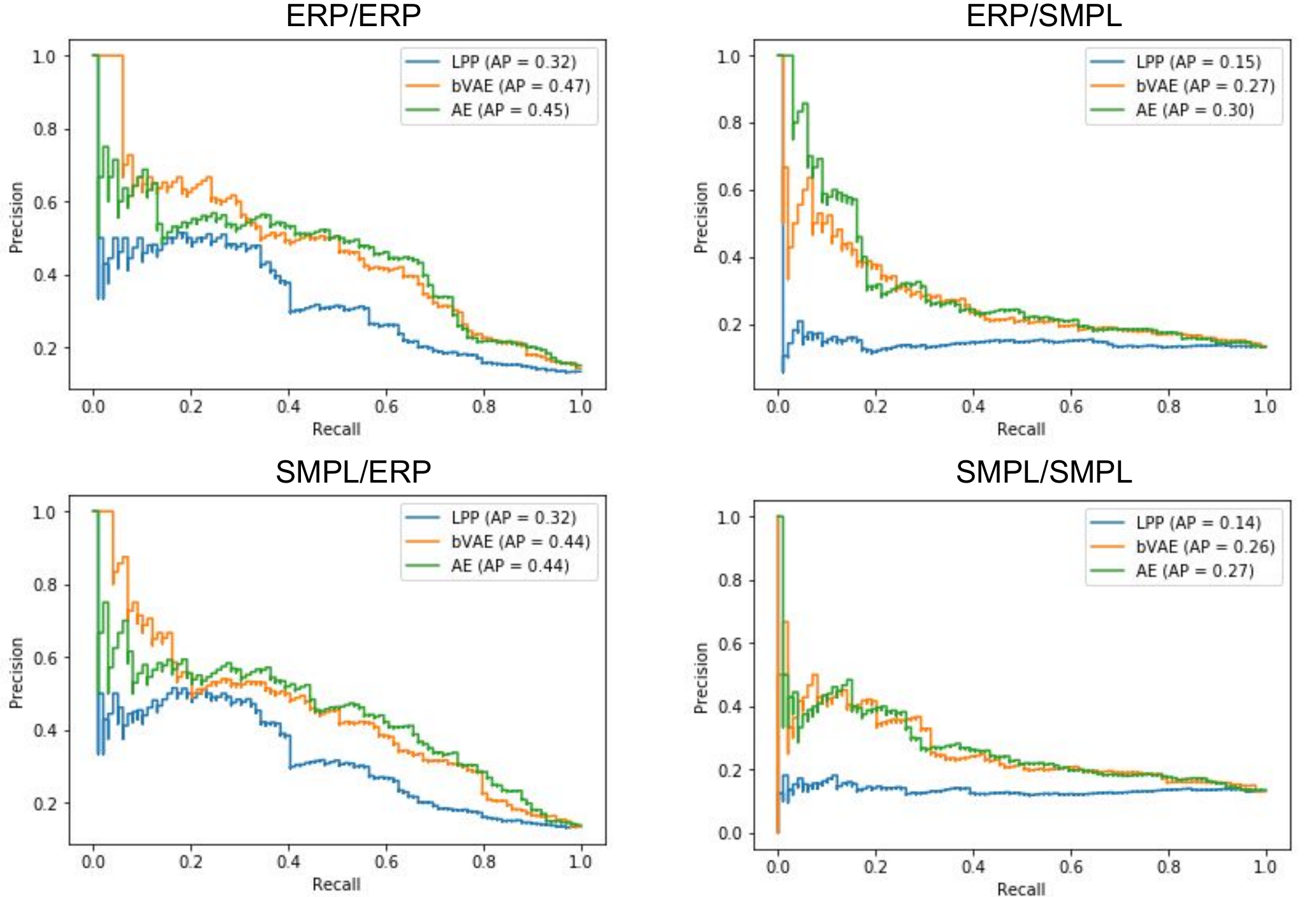}  
  \caption{ Precision-recall curves for balanced classification accuracy calculated as the average of precision and recall for L2 regularised logistic regression. ERP - averaged trajectories, SMPL - single sampled EEG trajectories. LPP - canonical late positive potential baseline representation. Models trained on single EEG trajectories (SMPL) are tested on different single EEG trajectories in the SMPL/SMPL train/test case.\color{black}  }
  \label{fig_precision_recall}
\end{wrapfigure}

\paragraph{Deep representation learning improves classification accuracy.} 
We evaluated the classification accuracy for predicting participant age, gender, study site, and the presence or absence of two clinical labels: depression and Axis 1. We first obtained the canonical LPP representations, and applied Support Vector Machine (SVM), Random Forest (RF), Logistic Regression (LR) and Linear Discriminant Analysis (LDA) to obtain balanced classification accuracy. Table~\ref{tbl_classification_results_baselines} and Figure~\ref{fig_precision_recall_lpp}  show that LR produced some of the best results overall for the baseline, hence we report only LR results for the $\beta$-VAE and AE representations in the main text (see Tables~\ref{tbl_classification_results_bvae}-\ref{tbl_classification_results_ae} and Figures~\ref{fig_precision_recall_bvae}-\ref{fig_precision_recall_ae} for the other classifiers applied to the $\beta$-VAE and AE representations).  Figures~\ref{fig_classification_results}-\ref{fig_precision_recall} ERP/ERP demonstrate that representations extracted through $\beta$-VAE and AE pretraining resulted in higher overall classification accuracy than those obtained through the baseline canonical LPP pipeline (see also Table~\ref{tbl_classification_results} ERP/ERP train/test cells). \color{black} This effect holds across all classification tasks, including depression and Axis 1 diagnoses. Furthermore, the pattern of classification results is in line with what might have been expected by the expert clinicians, whereby age and depression classification accuracy is significantly higher than chance, while gender and site are harder to decode from the EEG signal. Indeed, the only way to differentiate between study sites is by picking up on the noise artifacts produced by the differences in the ambient environment and other factors listed in Sec.~\ref{sec_intro}, which are hard to detect.

A similar pattern holds when all the models are trained on single EEG trials instead of ERPs (SMPL/ERP train/test in Figure~\ref{fig_classification_results}, see also Table~\ref{tbl_classification_results}). While the maximum classification accuracy tends to drop for all models when trained from the noisier single EEG trials rather than ERPs, the classification accuracy obtained from $\beta$-VAE and AE representations is still higher than that obtained from the LPP baseline.

\paragraph{Classification based on deep representations generalises better to single EEG trials.} One possible clinical application of the proposed classification pipeline is to enable online diagnosis recommendations from single EEG trials. Hence, we tested how well the classification accuracy of the different representations generalises to novel single EEG trial trajectories (*/SMPL in bars Figure~\ref{fig_classification_results}, see also SMPL columns in Tbl.~\ref{tbl_classification_results}). As expected, the performance tends to drop on average when classifying the noisy single EEG trajectories, regardless whether the representations were pre-trained using ERPs (ERP rows) or single EEG samples (SMPL rows, note that we tested the models with different single EEG samples to those used for training). However, the classification accuracy is still significantly higher for deep representations compared to the LPP baseline. This suggests that replacing the more manual canonical LPP pipeline with deep representation learning can allow for both better training data efficiency and a reduction in time that the (potentially vulnerable) participants have to spend in the lab by up to 37x, which is the average number of trials per condition that made it through EEG pre-processing in our dataset and were used for generating ERPs.

\paragraph{Deep representation learning reconstructs ERPs from single EEG trials.} Since $\beta$-VAE and AE had good classification accuracy when presented with single EEG samples, we tested whether they could also reconstruct ERPs from single EEG trials. Figure~\ref{fig_sample_reconstructions} shows that this is indeed the case---reconstructions produced by the pre-trained models from single noisy EEG samples look remarkably similar to those produced from the corresponding ERPs (also see Figs.~\ref{fig_sample_recs_1}-\ref{fig_sample_recs_3} for more examples). Note that this only holds true for models trained using ERP data and not those trained on single EEG samples.

\paragraph{Disentangled representations allow for interpretable classification.} To obtain SCAN classification results, we inferred the posterior $q_{\phi}(\mathbf{z}_x|\mathbf{x})$ of a pre-trained $\beta$-VAE in response to an EEG ``image'' $\mathbf{x}$. We then used the pre-trained SCAN decoder to obtain the reconstructed label logits $p_{\gamma}(\mathbf{y} | \mu(\mathbf{z}_x))$ using the $\beta$-VAE posterior mean as input (Fig.~\ref{fig_model}, red path). Finally, we applied softmax over the produced logits to obtain the predicted 5-hot label for the EEG ``image''. When SCAN was trained on top of a well disentangled $\beta$-VAE, it was able to outperform the canonical LPP baseline in terms of classification accuracy (see Figure~\ref{fig_classification_results} and Table~\ref{tbl_classification_results}, SCAN+$\beta$-VAE). This is despite the fact that SCAN was not developed with the classification goal in mind. Note, however, that SCAN can only work when trained with disentangled representations. Indeed, SCAN classification performance was not distinguishable from chance when trained on top of entangled AE representations (see Figure~\ref{fig_classification_results} and Table~\ref{tbl_classification_results}, SCAN+AE). To further confirm the role of disentanglement, we calculated Spearman correlation between the quality of disentanglement as measured by the UDR score and the final SCAN classification accuracy. Table~\ref{tbl_correlation_results} shows significant correlation for age, depression and Axis 1 diagnoses, suggesting that on average these factors were classified better if representations were more disentangled. The same, however, is not the case for gender or study site. This implies that some of the better disentangled models did not contain information necessary for classifying these factors. Such information loss is in fact a known trade-off of $\beta$-VAE training, with more disentangled $\beta$-VAE models often compromising on the informativeness of the learnt representation due to the increased compression pressure induced by the higher $\beta$ values necessary to achieve disentanglement \citep{higgins2017beta, duan2019unsupervised}.

When trained on top of well disentangled $\beta$-VAE models, SCAN decisions were not only accurate, but they were also based on a small number of disentangled dimensions. This is unlike the LR classifier, which obtained average sparsity of just 3.25\% compared to the 87.5\% for SCAN, even when L1 regularisation was used. Furthermore, the disentangled dimensions used by SCAN were interpretable, hence making SCAN classification decisions amenable to post-hoc analysis. Figure~\ref{fig_traversals}B visualises the inferred SCAN posterior when presented with 1-hot labels corresponding to the male or female gender, one of the four study sites, and the presence or absence of depression and Axis 1 diagnoses. In most cases, SCAN was able to associate the label with a single latent dimension in the pre-trained $\beta$-VAE (e.g., gender is represented by latent dimension $z_5$), and the different values of the same class label corresponded to disjoint narrow Gaussians on those latents (e.g., male gender is represented with a Gaussian  $\mathcal{N}(0.89,\ 0.49)$, while female gender is represented by a Gaussian $\mathcal{N}(-0.82,\ 0.52)$, both on $z_5$). We can also visualise what the different $\beta$-VAE latents have learnt to represent by plotting their traversals as shown in Figure~\ref{fig_traversals}A (see more in Figs.~\ref{fig_traversals_sup_1}-\ref{fig_traversals_sup_3}). 

Finally, we can also sample from the SCAN posterior $\hat{\mathbf{z}}_y \sim \mathcal{N}(\mu(\mathbf{z}_y|\mathbf{y}), \sigma(\mathbf{z}_y|\mathbf{y}))$ and reconstruct the resulting ERPs using the $\beta$-VAE decoder $p_{\theta}(\mathbf{x} | \hat{\mathbf{z}}_y)$ (Fig.~\ref{fig_model}, purple path). Such samples for the depression label are visualised in Figure~\ref{fig_traversals}C. It is clear that the reconstructed ERP samples corresponding to participants with a positive depression diagnosis contain no difference between the positive and neutral trials (overlapping blue and red lines), while those corresponding to participants without a positive depression diagnosis do have an obvious gap between the responses to positive and neutral trials. Hence, we were able to ``re-discover'' an objective biomarker for the symptom of anhedonia (and depression generally), thus opening up the potential for discovering new biomarkers hidden in EEG data in the future.
\section{Conclusions}

\begin{table}[t!]
\vskip -0.3in
\begin{center}
\begin{footnotesize}
\begin{tabular}{cc|cc|cc|cc|cc}
\toprule
\multicolumn{2}{c|}{Age} & \multicolumn{2}{c|}{Gender} & \multicolumn{2}{c|}{Site} & \multicolumn{2}{c|}{Depression} & \multicolumn{2}{c}{Axis 1 } \\
\textit{r}  & \textit{p} & \textit{r} & \textit{p} & \textit{r} & \textit{p} & \textit{r} & \textit{p} & \textit{r} & \textit{p} \\
\midrule
 0.28 & .01 & -0.1 & .31 & 0.19 & .06 & 0.33 & <.001 & 0.31 & <.001 \\
\bottomrule
\end{tabular}
\caption{Spearman correlation between disentanglement quality (as approximated by UDR score) of trained $\beta$-VAE models ($n$=100) and balanced classification accuracy from their corresponding SCAN models.} 
\label{tbl_correlation_results}
\end{footnotesize}
\end{center}
\vskip -0.2in
\end{table}

This work provides the first evidence that disentanglement-focused representation learning and the resulting models are powerful measurement tools with immediately applicability to applied and basic research for clinical psychology and electrophysiology. In particular, we have demonstrated that disentangled representation learning can be successfully applied to EEG data, resulting in representations that can be successfully re-used to predict multiple clinical factors through fast supervised re-mapping, outperforming a hand-engineered baseline typical for the field. Furthermore, our method recovers a representation that can be used to automatically extract denoised Event Related Potentials (ERPs) from novel, single EEG trajectories. Finally, single factors of the learned disentangled representations often correspond to meaningful markers of clinical factors (as automatically detected by SCAN), allowing for human interpretability and potentially providing novel insight into new biomarkers and the neurofunctional alterations underlying mental disorders.
\FloatBarrier


\subsubsection*{Acknowledgments}
We thank Sarah Laszlo, Gabriella Levine, Phil Watson, Obi Felten, Mustafa Ispir, Edward De Brouwer, the Amber team, Nader Amir and the Center for Understanding and Treating Anxiety, Dr. Kristen Schmidt, Alec Bruchnak, and Nicholas Santopetro.

\bibliography{iclr2021_conference}
\bibliographystyle{iclr2021_conference}

\newpage
\appendix
\section{Appendix}
\beginsupplement

\subsection{Experimental methods}
\subsubsection{Participants}
\label{app_sec_participants}
Both the participant and a legal guardian (where applicable) signed an informed consent form approved by their local university Institutional Review Board (IRB) and all procedures were conducted in line with the Declaration of Helsinki.  Presentation \textregistered (Neurobehavioural Systems Inc, \href{www.neurobs.com}{www.neurobs.com}) was used for stimulus presentation.\color{black} 

\begin{table}[h!]
\begin{center}
\begin{scriptsize}
\begin{tabular}{ccccrr}
\toprule
 & &  &   &   &\\
Site & AgeGroup & Gender & Diagnosis & Axis 1 & Depression   \\
\midrule
Site 1 & adult & F & 0.0 &  35 & 35\\ 
 & &  & 1.0 &  59 & 59\\
 & & M & 0.0 &  11 & 11\\
 & &  & 1.0 &  12 & 12\\
Site 2 & child & F & 0.0 &  72 & 95 \\ 
 & &  & 1.0 &  31 & 8 \\
 & & M & 0.0 & 101 & 119 \\
 & &  & 1.0 &  19 & 1 \\
Site 3 & child & F & 0.0 &  82 & 128\\ 
 & &  & 1.0 &  57 & 11\\
 & & M & 0.0 &  98 & 150\\
 & &  & 1.0 &  54 & 2\\
Site 4 & child & F & 0.0 &  41 & 58\\ 
 & &  & 1.0 &  21 & 4\\
 & & M & 0.0 &  45 & 63\\
 & &  & 1.0 &  20 & 2\\
\bottomrule
\end{tabular}
\caption{Participant label distribution}
\label{tbl_participant_demos}
\end{scriptsize}
\end{center}
\end{table}

\subsubsection{EEG preprocessing}
\label{app_eeg_preprocessing}
EEG was recorded from a 34-channel ActiChamp system (BioSemi, Amsterdam, Netherlands), positioned according to the 10/20 system. The data was recorded at 250~Hz and referenced to electrode Cz. To monitor eye blinks, VEOG and HEOG electrodes were applied above, below and to the outer canthi of both eyes. The raw EEG data was digitized at a 24-bit resolution using a low-pass 5th order sinc filter with a half-power cutoff at 102.4~Hz.

Offline, the continuous EEG data were re-referenced to the mastoids and bandpass filtered using a 2nd order Butterworth high-pass filter with a 3~dB cut-off at 0.1~Hz and a 12th order Butterworth low-pass filter with a 3~dB cut-off at 30~Hz. Eye-movements were detected and removed from the continuous data using ICA \citep{gratton1983new}.  In this process, the first ICA component was always targeted as the to-be-corrected blink artifact (allowing for a fully automated ICA pipeline). EEG from three electrodes were chosen to be included in the analysis---Fz, Cz, Pz---because (1) the LPP is traditionally indexed from centro-parietal midline electrodes and (2) it was desirable for the project to test an EEG system that maximizes speed of application and ease of use. The full preprocessing pipeline was built with the MNE \citep{gramfort2014mne} software package in Python.\color{black}

The final step in the pre-processing pipeline was to exclude anomalous trials (EEG segments) from each participants data. In order of application, a raw maximum amplitude threshold was applied at \SI{\pm.005}{\volt}, the data were normalized per-participant to zero mean and unit variance and a standard deviation threshold was applied at $|\sigma|>=5$. Finally, standard deviation ``transients'' (i.e., deflections in the EEG time series between $t$ and $t+1$ where $|\sigma|>=3$) were identified and the associated trials were removed. Overall 758 trajectories were used to train the models. \color{black}

\subsection{Models}

\subsubsection{Variational Autoencoder}
\label{app_vae}
A Variational Autoencoder (VAE) \citep{rezende2014stochastic,kingma2013auto} is a generative model that aims to learn a joint distribution $p(\mathbf{x}, \mathbf{z})$ of input data $\mathbf{x}$ and a set of latent variables $\mathbf{z}$ by learning to maximize the Evidence Lower BOund (ELBO) on the data distribution. The neural network implementation consists of an inference network (equivalent to the AE encoder), that takes inputs $\mathbf{x}$ and parameterizes the posterior distribution $q(\mathbf{z}|\mathbf{x})$, and a generative network (equivalent to the AE decoder) that takes a sample from the inferred posterior distribution $\hat{\mathbf{z}} \sim \mathcal{N}(\mu(\mathbf{z}|\mathbf{x}), \sigma(\mathbf{z}|\mathbf{x}))$ and attempts to reconstruct the original image. The model is trained through a two-part loss objective:

\begin{equation*}
   \mathcal{L}_{VAE} = \mathbb{E}_{p(\mathbf{x})} [\ \mathbb{E}_{q_{\phi}(\mathbf{z}|\mathbf{x})} [\log\ p_{\theta}(\mathbf{x} | \mathbf{z})] - KL(q_{\phi}(\mathbf{z}|\mathbf{x})\ ||\ p(\mathbf{z}))\ ]
\end{equation*}

where $p(\mathbf{x})$ is the probability of the input data, $q(\mathbf{z}|\mathbf{x})$ is the learnt posterior over the latent units given the data, $p(\mathbf{z})$ is the unit Gaussian prior with a diagonal covariance matrix $\mathcal{N}(\mathbf{0}, \mathbb{I})$, and $\phi$ and $\theta$ are the parameters of the inference (encoder) and generative (decoder) networks respectively. The encoder parametrised a Gaussian distribution with a diagonal covariance matrix, while the decoder parametrised a Bernoulli distribution.


\subsection{Classifiers}
\label{app_classifiers}
In each case, we extracted representations for the full dataset, and used them as input into the baseline classifiers, using the sklearn package implementation with default parameters, unless specified otherwise. In particular, given that the label data was often significantly unbalanced, we used the ``balanced'' class weight option where appropriate, which adjusted the weight for each dataset sample to be inversely proportional to its corresponding class frequency according to $w_i = N/n_{y}$, where $w_i$ is the weight for sample $i$, $n_{y}$ is the number of samples in the dataset with the same class label as $y_i$, and $N$ is the total number of datapoints. We used 5-fold cross-validation to choose the best hyperparameters to avoid overfitting where appropriate. For LR we used 500 maximum iterations, and both L1 and L2 regularisation. For RF we used depth 2 (larger values did not improve performance).

\subsection{Unsupervised Disentanglement Ranking}
\label{app_udr}
In order to quantify the quality of disentanglement achieved by the trained $\beta$-VAE models, we applied the recently proposed Unsupervised Disentanglement Ranking (UDR) score \citep{duan2019unsupervised}. UDR is the only known method for measuring the quality of disentanglement in VAEs without access to the ground truth attribute labels, which were not available for the EEG data. UDR works by performing pairwise comparisons between the representations learned by models trained using the same hyperparameter setting but with different seeds. For each trained $\beta$-VAE model we performed nine pairwise comparisons with all the other models trained with the same $\beta$ value and calculated the corresponding UDR$_{ij}$ score, where $i$ and $j$ index the two $\beta$-VAE models. Each UDR$_{ij}$ score is calculated by computing the similarity matrix $R_{ij}$, where each entry is the Spearman correlation between the responses of individual latent units of the two models across the same data. The absolute value of the similarity matrix is then taken $|R_{ij}|$ and the final score for each pair of models is calculated according to: 

\begin{equation*}
\label{rel_strength}
\frac{1}{d_a+d_b}\left [ \sum_b \frac{r_a^2 * I_{KL}(b)}{\sum_a R (a, b)} + \sum_a \frac{r_b^2 * I_{KL}(a)}{\sum_b R (a, b)} \right ] 
\end{equation*}

where $a$ and $b$ index into the latent units of models $i$ and $j$ respectively, $r_a = \max_a R (a, b)$ and $r_b = \max_b R (a, b)$. $I_{KL}$ indicate the ``informative'' latent units within each model, operationalised as units with $KL>0.01$ from the unit Gaussian prior, and $d$ is the number of such latent units. The final score for model $i$ is calculated by taking the median of UDR$_{ij}$ across all $j$.

\begin{figure}[h!]
 \centering
 \includegraphics[width=0.5\textwidth]{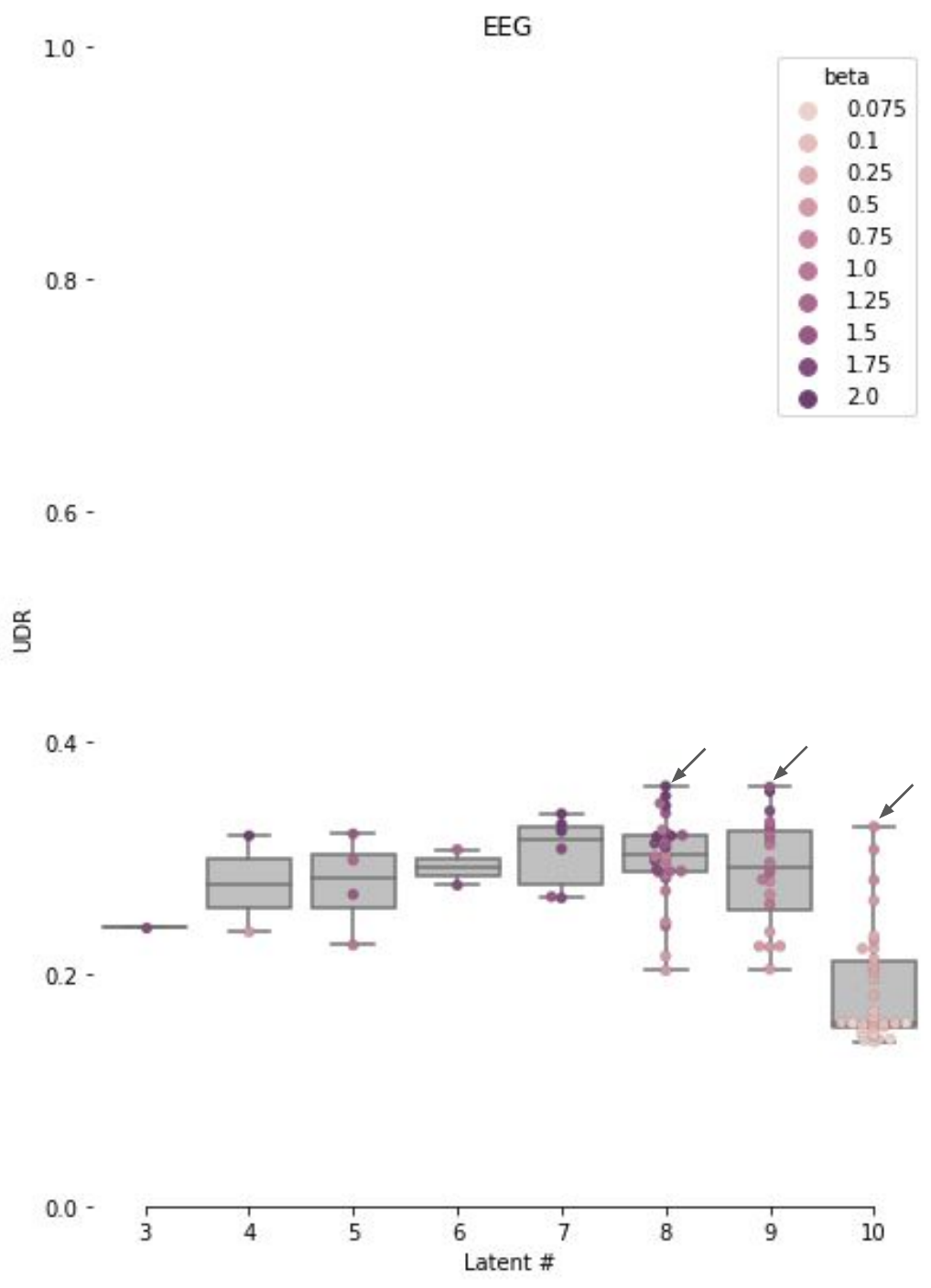}
 \caption{Distribution of UDR scores against the final number of ``informative'' latents across the hyperparameter search for $\beta$-VAE. A latent is ``informative'' if the distance of its posterior against the unit Gaussian prior is high $KL>0.01$, otherwise the latent carries no information and is effectively switched off. Three models whose latent traversals are plotted in Figs.~\ref{fig_traversals_sup_1}-\ref{fig_traversals_sup_3} are indicated by arrows.}
 \label{fig_udr_scores}
\end{figure}

\begin{table}[t!]
\begin{center}
\begin{tiny}
\begin{tabular}{lll|cc|cc|cc|cc|cc||cc}
\toprule
\multicolumn{3}{l|}{} & \multicolumn{2}{c|}{Age} & \multicolumn{2}{c|}{Gender} & \multicolumn{2}{c|}{Site} & \multicolumn{2}{c|}{Depression} & \multicolumn{2}{c||}{Axis 1 } & \multicolumn{2}{c}{Overall} \\
& Train $\downarrow$ & Test $\rightarrow$ & ERP & SMPL & ERP & SMPL & ERP & SMPL & ERP & SMPL & ERP & SMPL & ERP & SMPL \\
\midrule
\multicolumn{3}{l|}{Chance}  & \multicolumn{2}{c|}{0.50} &\multicolumn{2}{c|}{0.50} &\multicolumn{2}{c|}{0.25} &\multicolumn{2}{c|}{0.50} & \multicolumn{2}{c||}{0.50} & \multicolumn{2}{c}{0.45} \\
\midrule
LR & ERP & \multirow{2}{*}{LPP} & 0.74 & \textbf{0.56} & 0.52 & 0.51 & 0.38 & \textbf{0.30} & \textbf{0.64} & \textbf{0.56} & 0.52 & 0.52 & 0.56 & \textbf{0.49} \\
LR & SMPL &  & 0.69 & 0.54 & 0.52 & \textbf{0.52} & 0.36 & \textbf{0.30} & 0.61 & 0.54 & 0.52 & 0.52 & 0.54 & \textbf{0.48} \\
\midrule
RF & ERP &  \multirow{2}{*}{LPP} & \textbf{0.80} & 0.54 & \textbf{0.59} & \textbf{0.53} & 0.34 & 0.21 & \textbf{0.64} & 0.47 & \textbf{0.63} & 0.53 & \textbf{0.60} & 0.45 \\
RF & SMPL & & 0.46 & 0.46 & 0.52 & \textbf{0.52} & 0.24 & 0.20 & 0.47 & 0.47 & 0.54 & 0.41 & 0.44 & 0.41 \\
\midrule
LDA & ERP & \multirow{2}{*}{LPP}   & 0.78 & 0.54 & 0.51 & 0.48 & 0.32 & 0.28 & 0.62 & 0.50 & 0.52 & 0.54 & 0.55 & 0.47 \\
LDA & SMPL &  & 0.46 & 0.46 & 0.51 & 0.51 & 0.34 & 0.27 & 0.47 & 0.47 & 0.58 & \textbf{0.66} & 0.47 & 0.47 \\
\midrule
SVM & ERP &  \multirow{2}{*}{LPP} & 0.76 & 0.54 & \textbf{0.58} & 0.51 & \textbf{0.41} & 0.27 & \textbf{0.65} & \textbf{0.55} & 0.57 & 0.51 & \textbf{0.59} & \textbf{0.48} \\
SVM & SMPL & & 0.53 & \textbf{0.55} & 0.52 & \textbf{0.53} & 0.28 & 0.27 & 0.61 & 0.52 & 0.52 & 0.49 & 0.49 & 0.47 \\

\bottomrule
\end{tabular}
\caption{Balanced classification accuracy calculated as the average of precision and recall. ERP - averaged trajectories, SMPL - single sampled EEG trajectories. LPP - canonical late positive potential baseline representation. LR - L2 regularised logistic regression, RF - random forest, LDA - linear discriminant analysis, SVM - support vector machine. Models trained on single EEG trajectories (SMPL) are tested on different single EEG trajectories in the SMPL/SMPL train/test case. Average balanced classification accuracy across all conditions: LR: 0.52, RF: 0.48, LDA: 0.49, SVM: 0.51}
\label{tbl_classification_results_baselines}
\end{tiny}
\end{center}
\end{table}

\begin{table}[t!]
\begin{center}
\begin{tiny}
\begin{tabular}{lll|cc|cc|cc|cc|cc||cc}
\toprule
\multicolumn{3}{l|}{} & \multicolumn{2}{c|}{Age} & \multicolumn{2}{c|}{Gender} & \multicolumn{2}{c|}{Site} & \multicolumn{2}{c|}{Depression} & \multicolumn{2}{c||}{Axis 1 } & \multicolumn{2}{c}{Overall} \\
& Train $\downarrow$ & Test $\rightarrow$ & ERP & SMPL & ERP & SMPL & ERP & SMPL & ERP & SMPL & ERP & SMPL & ERP & SMPL \\
\midrule
\multicolumn{3}{l|}{Chance}  & \multicolumn{2}{c|}{0.50} &\multicolumn{2}{c|}{0.50} &\multicolumn{2}{c|}{0.25} &\multicolumn{2}{c|}{0.50} & \multicolumn{2}{c||}{0.50} & \multicolumn{2}{c}{0.45} \\
\midrule

LR & ERP & \multirow{2}{*}{$\beta$-VAE} & 0.85 & 0.62 & 0.66 & 0.53 & 0.51 & 0.34 & 0.68 & 0.59 & 0.60 & 0.55 & 0.66 & \textbf{0.53} \\
LR & SMPL &  & 0.76 & 0.65 & 0.58 & 0.53 & 0.40 & \textbf{0.36} & 0.67 & 0.58 & 0.56 & 0.56 & 0.59 & \textbf{0.54} \\
\midrule
RF & ERP &  \multirow{2}{*}{$\beta$-VAE} & 0.75 & \textbf{0.56} & 0.75 & 0.56 & 0.45 & 0.29 & 0.47 & 0.47 & 0.66 & 0.54 & 0.62 & 0.48 \\
RF & SMPL & &  0.78 & \textbf{0.72} & 0.65 & 0.55 & 0.43 & 0.29 & 0.70 & 0.47 & 0.65 & \textbf{0.66} & 0.64 & \textbf{0.54} \\
\midrule
LDA & ERP & \multirow{2}{*}{$\beta$-VAE} & 0.86 & 0.63 & 0.66 & 0.54 & 0.46 & 0.31 & 0.68 & \textbf{0.63} & 0.58 & 0.54 & 0.65 & \textbf{0.53} \\
LDA & SMPL &  &  0.82 & 0.61 & 0.63 & 0.54 & 0.50 & 0.31 & 0.69 & 0.52 & 0.56 & 0.53 & 0.64 & 0.50 \\
\midrule
SVM & ERP &  \multirow{2}{*}{$\beta$-VAE} & \textbf{0.93} & 0.61 & \textbf{0.78} & 0.54 & \textbf{0.71} & 0.33 & \textbf{0.83} & 0.59 & \textbf{0.74} & 0.57 & \textbf{0.80} & \textbf{0.53} \\
SVM & SMPL & &  0.88 & 0.64 & 0.71 & \textbf{0.56} & 0.60 & 0.34 & 0.80 & 0.57 & 0.69 & 0.53 & 0.74 & \textbf{0.53} \\

\bottomrule
\end{tabular}
\caption{Balanced classification accuracy calculated as the average of precision and recall. ERP - averaged trajectories, SMPL - single sampled EEG trajectories. LPP - canonical late positive potential baseline representation. LR - L2 regularised logistic regression, RF - random forest, LDA - linear discriminant analysis, SVM - support vector machine. Models trained on single EEG trajectories (SMPL) are tested on different single EEG trajectories in the SMPL/SMPL train/test case. Average balanced classification accuracy across all conditions: LR: 0.58, RF: 0.57, LDA: 0.58, SVM: 0.65}
\label{tbl_classification_results_bvae}
\end{tiny}
\end{center}
\end{table}

\begin{table}[t!]
\begin{center}
\begin{tiny}
\begin{tabular}{lll|cc|cc|cc|cc|cc||cc}
\toprule
\multicolumn{3}{l|}{} & \multicolumn{2}{c|}{Age} & \multicolumn{2}{c|}{Gender} & \multicolumn{2}{c|}{Site} & \multicolumn{2}{c|}{Depression} & \multicolumn{2}{c||}{Axis 1 } & \multicolumn{2}{c}{Overall} \\
& Train $\downarrow$ & Test $\rightarrow$ & ERP & SMPL & ERP & SMPL & ERP & SMPL & ERP & SMPL & ERP & SMPL & ERP & SMPL \\
\midrule
\multicolumn{3}{l|}{Chance}  & \multicolumn{2}{c|}{0.50} &\multicolumn{2}{c|}{0.50} &\multicolumn{2}{c|}{0.25} &\multicolumn{2}{c|}{0.50} & \multicolumn{2}{c||}{0.50} & \multicolumn{2}{c}{0.45} \\
\midrule
LR & ERP & \multirow{2}{*}{AE}  & 0.85 & 0.62 & 0.68 & \textbf{0.57} & 0.48 & \textbf{0.37} & 0.69 & 0.59 & 0.60 & \textbf{0.56} & 0.66 & \textbf{0.54} \\
LR & SMPL &  & 0.76 & \textbf{0.65} & 0.63 & \textbf{0.57} & 0.46 & \textbf{0.36} & 0.66 & 0.58 & 0.59 & 0.54 & 0.62 & \textbf{0.54} \\
\midrule
RF & ERP &  \multirow{2}{*}{AE} &  0.72 & 0.46 & 0.71 & \textbf{0.57} & 0.44 & 0.29 & 0.47 & 0.47 & 0.66 & 0.41 & 0.60 & 0.44  \\
RF & SMPL & &  0.77 & \textbf{0.64} & 0.68 & \textbf{0.57} & 0.45 & 0.32 & 0.74 & 0.47 & 0.62 & 0.54 & 0.65 & 0.51  \\
\midrule
LDA & ERP & \multirow{2}{*}{AE} & 0.87 & 0.63 & 0.67 & \textbf{0.58} & 0.46 & 0.31 & 0.67 & \textbf{0.70} & 0.59 & 0.50 & 0.65 & \textbf{0.54} \\
LDA & SMPL &  & 0.86 & 0.61 & 0.67 & 0.56 & 0.46 & 0.33 & 0.70 & 0.62 & 0.58 & 0.54 & 0.65 & \textbf{0.53}  \\
\midrule
SVM & ERP &  \multirow{2}{*}{AE} & \textbf{0.92} & 0.63 & \textbf{0.80} & 0.56 & \textbf{0.69} & \textbf{0.36} & \textbf{0.82} & 0.60 & \textbf{0.76} & \textbf{0.55} & \textbf{0.80} & \textbf{0.54} \\
SVM & SMPL & &  0.89 & \textbf{0.65} & 0.75 & \textbf{0.57} & 0.60 & \textbf{0.36} & 0.80 & 0.58 & 0.69 & 0.54 & 0.75 & \textbf{0.54}  \\
\bottomrule
\end{tabular}
\caption{Balanced classification accuracy calculated as the average of precision and recall. ERP - averaged trajectories, SMPL - single sampled EEG trajectories. LPP - canonical late positive potential baseline representation. LR - L2 regularised logistic regression, RF - random forest, LDA - linear discriminant analysis, SVM - support vector machine. Models trained on single EEG trajectories (SMPL) are tested on different single EEG trajectories in the SMPL/SMPL train/test case. Average balanced classification accuracy across all conditions: LR: 0.59, RF: 0.55, LDA: 0.60, SVM: 0.66}
\label{tbl_classification_results_ae}
\end{tiny}
\end{center}
\end{table}

\begin{figure}[t!]
 \includegraphics[width=1.\textwidth]{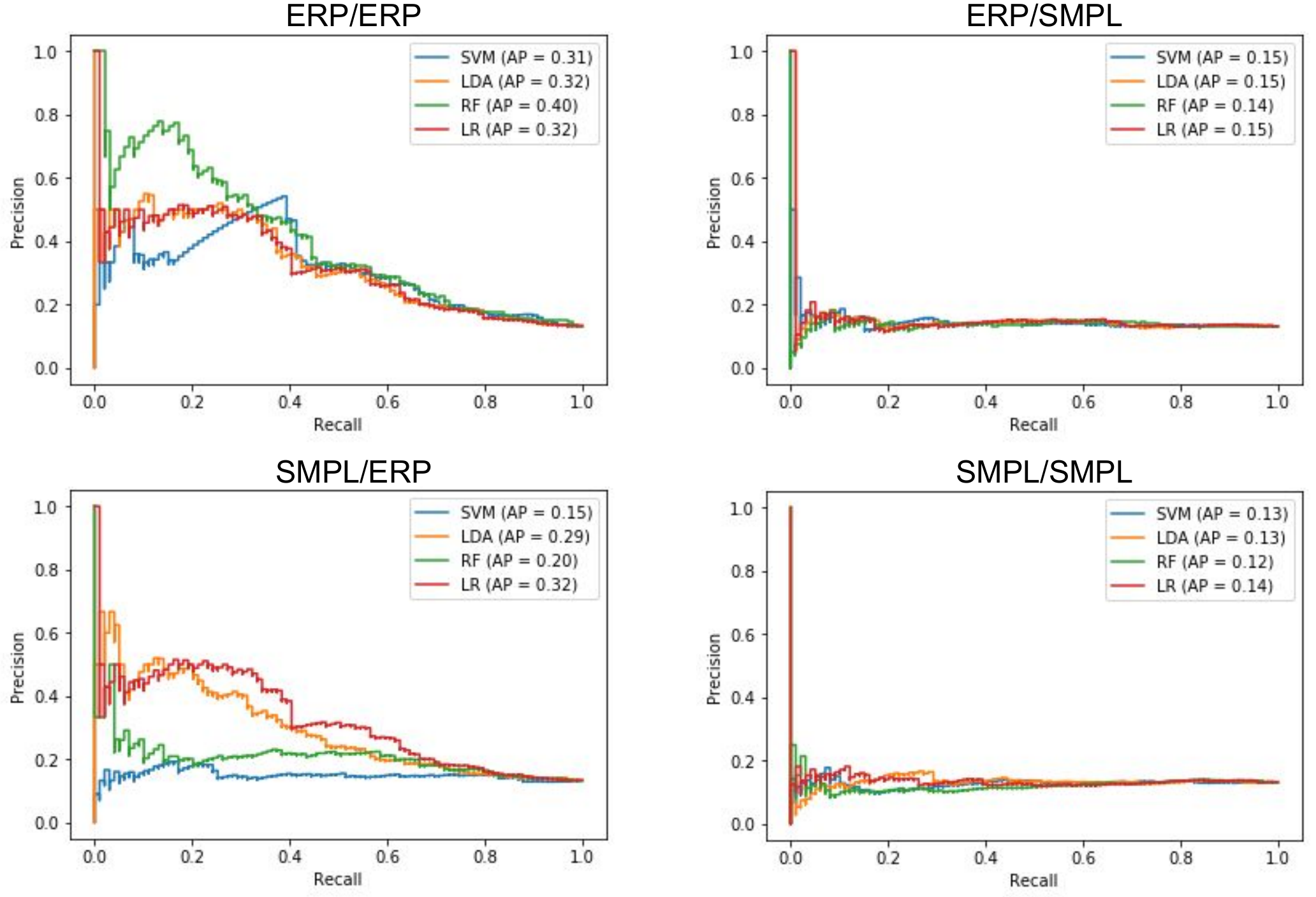} 
 \caption{Precision-recall curves for balanced classification accuracy calculated as the average of precision and recall for canonical late positive potential (LPP) baseline representation. LR - L2 regularised logistic regression, RF - random forest, LDA - linear discriminant analysis, SVM - support vector machine. }
 \label{fig_precision_recall_lpp}
\end{figure}

\begin{figure}[t!]
 \includegraphics[width=1.\textwidth]{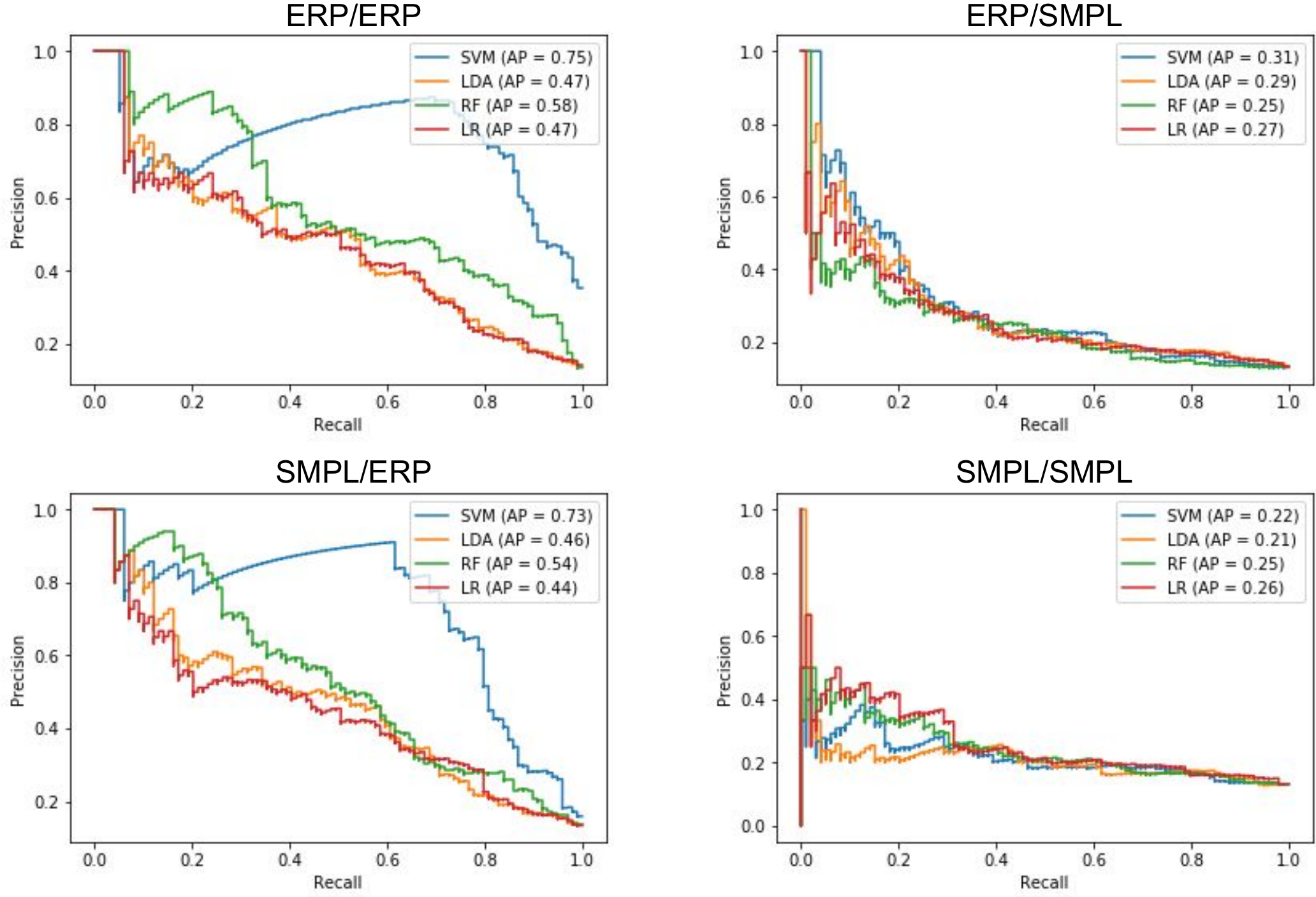} 
 \caption{Precision-recall curves for balanced classification accuracy calculated as the average of precision and recall for $\beta$-VAE. LR - L2 regularised logistic regression, RF - random forest, LDA - linear discriminant analysis, SVM - support vector machine. }
 \label{fig_precision_recall_bvae}
\end{figure}

\begin{figure}[t!]
 \includegraphics[width=1.\textwidth]{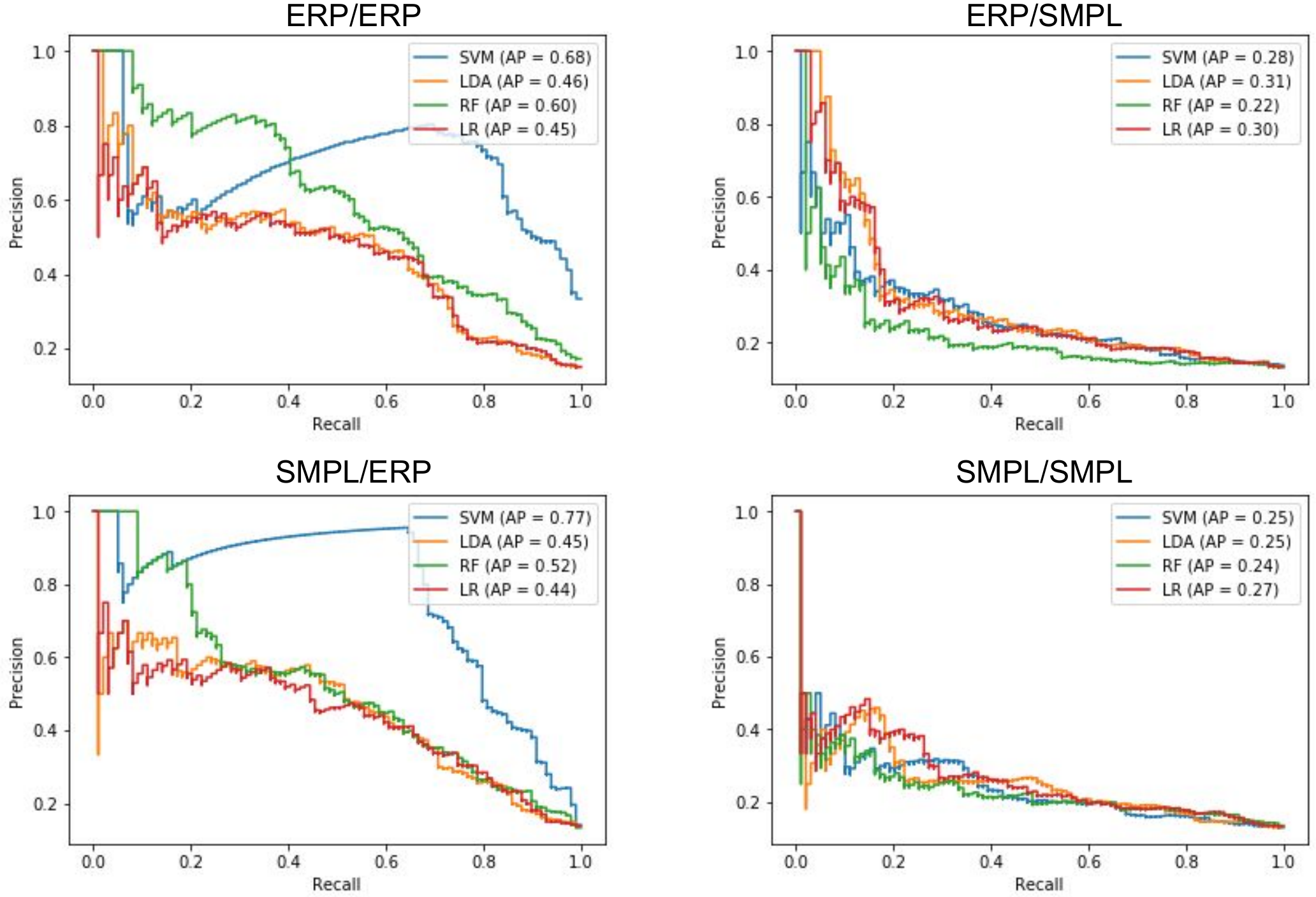} 
 \caption{Precision-recall curves for balanced classification accuracy calculated as the average of precision and recall for AE. LR - L2 regularised logistic regression, RF - random forest, LDA - linear discriminant analysis, SVM - support vector machine. }
 \label{fig_precision_recall_ae}
\end{figure}

\begin{table}[t!]
\vskip -0.3in
\begin{center}
\begin{tiny}
\begin{tabular}{lll|cc|cc|cc|cc|cc||cc}
\toprule
\multicolumn{3}{l|}{} & \multicolumn{2}{c|}{Age} & \multicolumn{2}{c|}{Gender} & \multicolumn{2}{c|}{Site} & \multicolumn{2}{c|}{Depression} & \multicolumn{2}{c||}{Axis 1 } & \multicolumn{2}{c}{Overall} \\
& Train $\downarrow$ & Test $\rightarrow$ & ERP & SMPL & ERP & SMPL & ERP & SMPL & ERP & SMPL & ERP & SMPL & ERP & SMPL \\

\midrule

\multicolumn{3}{l|}{Chance}  & \multicolumn{2}{c|}{0.50} &\multicolumn{2}{c|}{0.50} &\multicolumn{2}{c|}{0.25} &\multicolumn{2}{c|}{0.50} & \multicolumn{2}{c||}{0.50} & \multicolumn{2}{c}{0.45} \\

\midrule

LR & ERP & \multirow{2}{*}{LPP} & 0.74 & 0.56 & 0.52 & 0.51 & 0.38 & 0.30 & 0.64 & 0.56 & 0.52 & 0.52 & 0.56 & 0.49 \\
LR & SMPL &  & 0.69 & 0.54 & 0.52 & 0.52 & 0.36 & 0.30 & 0.61 & 0.54 & 0.52 & 0.52 & 0.54 & 0.48 \\

\midrule
\midrule
    
\multirow{2}{*}{LR} & \multirow{2}{*}{ERP} & $\beta$-VAE   & \textbf{0.85} & 0.62 & 0.66 & 0.53 & \textbf{0.51} & 0.34 & \textbf{0.68} & \textbf{0.59} & \textbf{0.60} & \textbf{0.55} & \textbf{0.66} & \textbf{0.53} \\
   &  & AE   & \textbf{0.85} & 0.62 & \textbf{0.68} & \textbf{0.57} & 0.48 & \textbf{0.37} & \textbf{0.69} & \textbf{0.59} & \textbf{0.60} & \textbf{0.56} & \textbf{0.66} & \textbf{0.54} \\

\multirow{2}{*}{LR} & \multirow{2}{*}{SMPL} & $\beta$-VAE  & 0.76 & \textbf{0.65} & 0.58 & 0.53 & 0.40 & \textbf{0.36} & 0.67 & \textbf{0.58} & 0.56 & \textbf{0.56} & 0.59 & \textbf{0.54} \\
  &  & AE & 0.76 & \textbf{0.65} & 0.63 & \textbf{0.57} & 0.46 & \textbf{0.36} & 0.66 & \textbf{0.58} & \textbf{0.59} & 0.54 & 0.62 & \textbf{0.54} \\  
 
\midrule

\multirow{2}{*}{SCAN} & \multirow{2}{*}{ERP} & $\beta$-VAE   & 0.78 & \textbf{0.64} & 0.58 & 0.50 & 0.39 & 0.30 & \textbf{0.68} & \textbf{0.59} & 0.54 & 0.52 & 0.59 & 0.51 \\
   & & AE   & 0.48 & 0.50 & 0.48 & 0.50 & 0.21 & 0.25 & 0.50 & 0.49 & 0.50 & 0.50 & 0.43 & 0.45 \\

\multirow{2}{*}{SCAN} & \multirow{2}{*}{SMPL} & $\beta$-VAE & 0.73 & 0.59 & 0.57 & 0.53 & 0.38 & 0.30 & 0.67 & 0.56 & 0.55 & 0.50 & 0.58 & 0.5 \\
    & & AE   & 0.41 & 0.45 & 0.43 & 0.46 & 0.16 & 0.20 & 0.46 & 0.49 & 0.53 & 0.49 & 0.40 & 0.42 \\

\bottomrule
\end{tabular}
\caption{Balanced classification accuracy calculated as the average of precision and recall $Acc = \left ( \frac{TP}{TP+FN} + \frac{TN}{TN+FP} \right ) / 2$. ERP - averaged trajectories, SMPL - single sampled EEG trials. LPP - canonical late positive potential baseline representation. LR - L2 regularised logistic regression (see Tbl.~\ref{tbl_classification_results_baselines} for other baseline classifiers). Models trained on single EEG trajectories (SMPL) are tested on different single EEG trajectories in the SMPL/SMPL train/test case.} 
\label{tbl_classification_results}
\end{tiny}
\end{center}
\vskip -0.2in
\end{table}

\begin{figure}[h!]
 \centering
 \includegraphics[width=1.\textwidth]{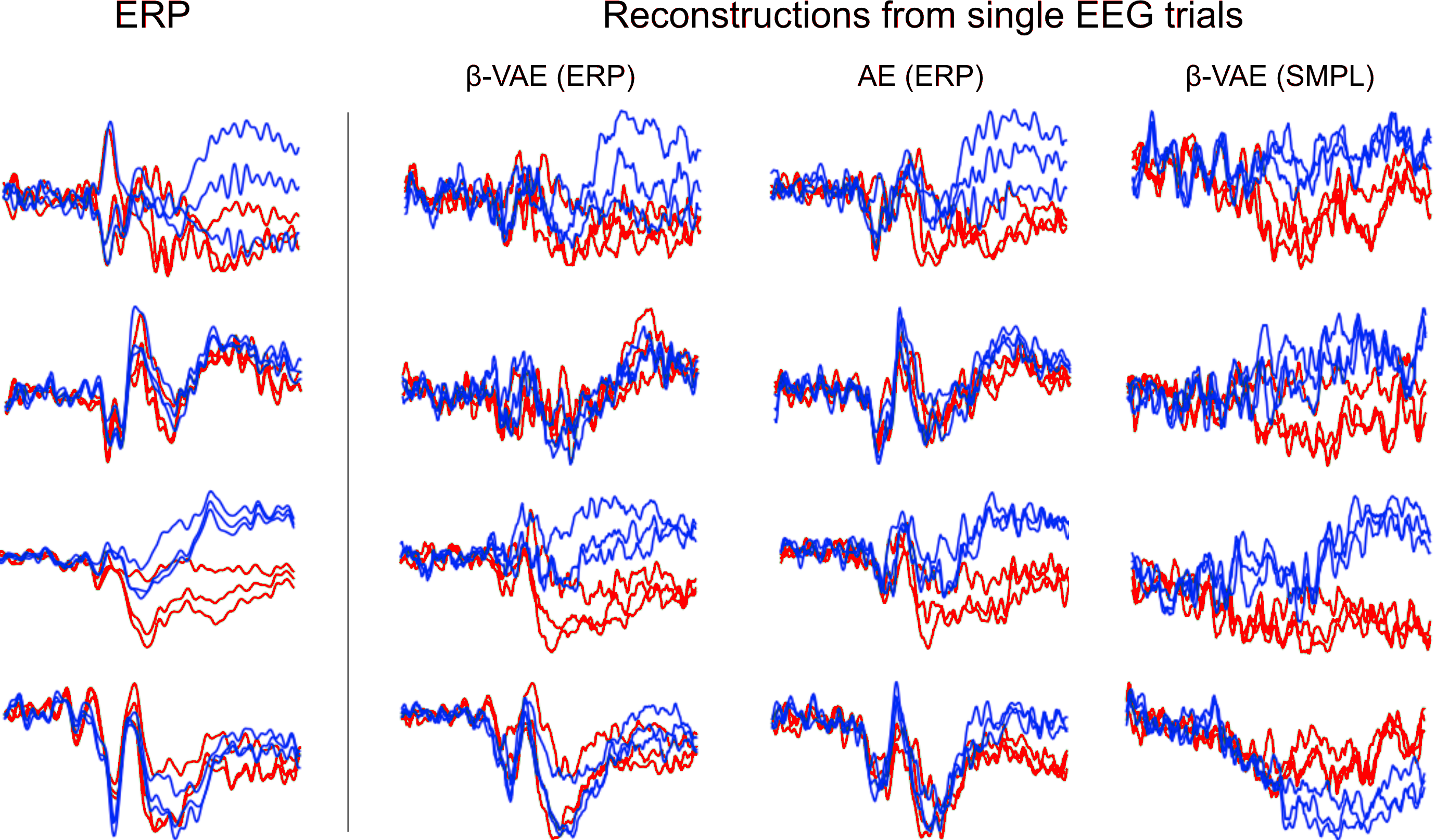}
 \caption{Reconstructions from single EEG trials using two disentangled $\beta$-VAEs pre-trained on either ERPs or single EEG sample trajectories (SMPL), as well as an AE pre-trained on ERPs (AE ERP). Models pre-trained on ERPs are able to reconstruct ERP-like trajectories even from single EEG samples, as demonstrated by the closer similarity of the $\beta$-VAEs (ERP) and AE (ERP) reconstructions to the ground truth ERPs (leftmost column). The ground truth ERPs were obtained by averaging on average 37 single EEG trials, including the single EEG trajectories that were used as inputs to the models.}
 \label{fig_sample_recs_1}
\end{figure}

\begin{figure}[h!]
 \centering
 \includegraphics[width=1.\textwidth]{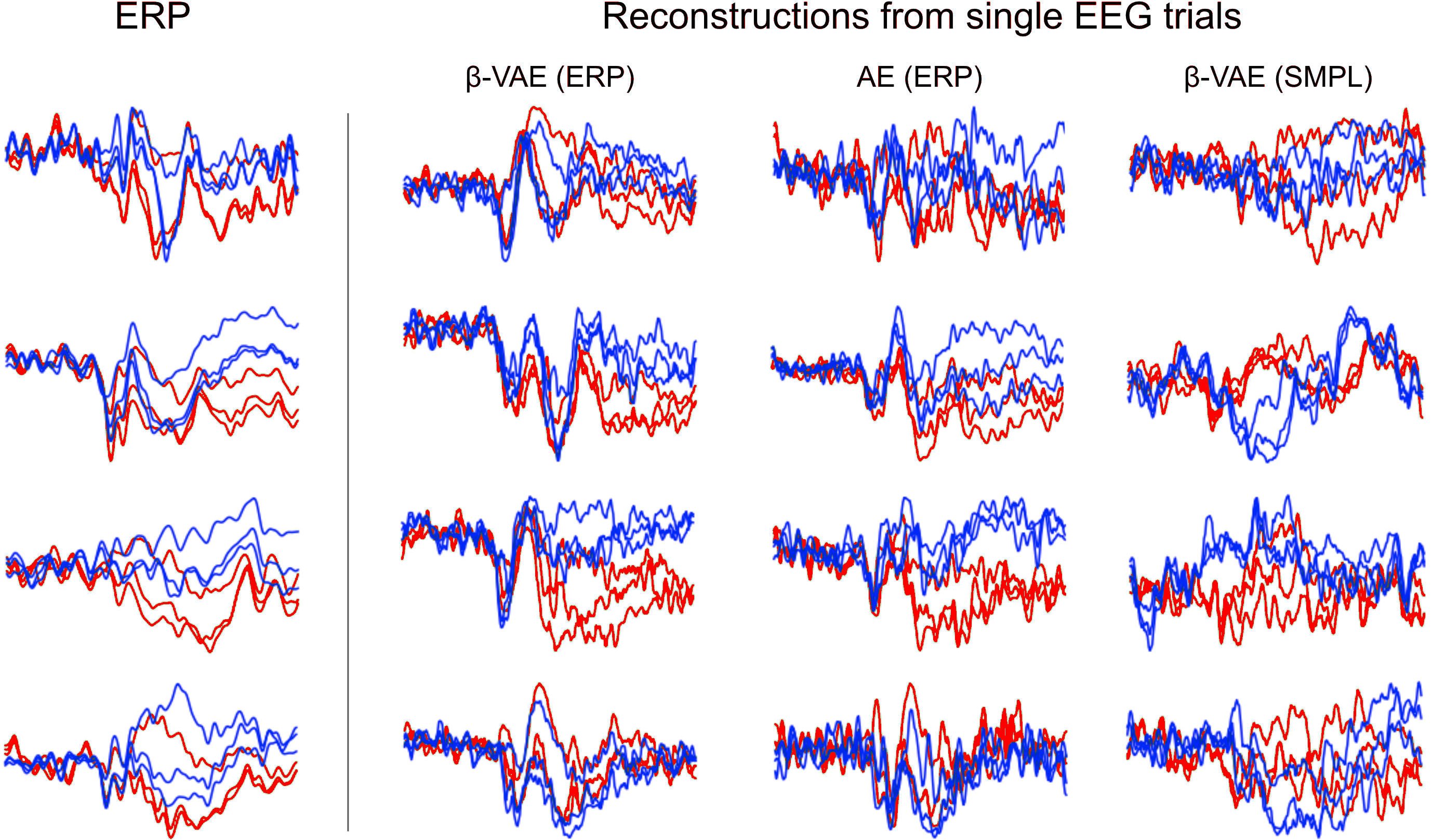}
 \caption{Reconstructions from single EEG trials using two disentangled $\beta$-VAEs pre-trained on either ERPs or single EEG sample trajectories (SMPL), as well as an AE pre-trained on ERPs (AE ERP). Models pre-trained on ERPs are able to reconstruct ERP-like trajectories even from single EEG samples, as demonstrated by the closer similarity of the $\beta$-VAEs (ERP) and AE (ERP) reconstructions to the ground truth ERPs (leftmost column). The ground truth ERPs were obtained by averaging on average 37 single EEG trials, including the single EEG trajectories that were used as inputs to the models.}
 \label{fig_sample_recs_2}
\end{figure}

\begin{figure}[h!]
 \centering
 \includegraphics[width=1.\textwidth]{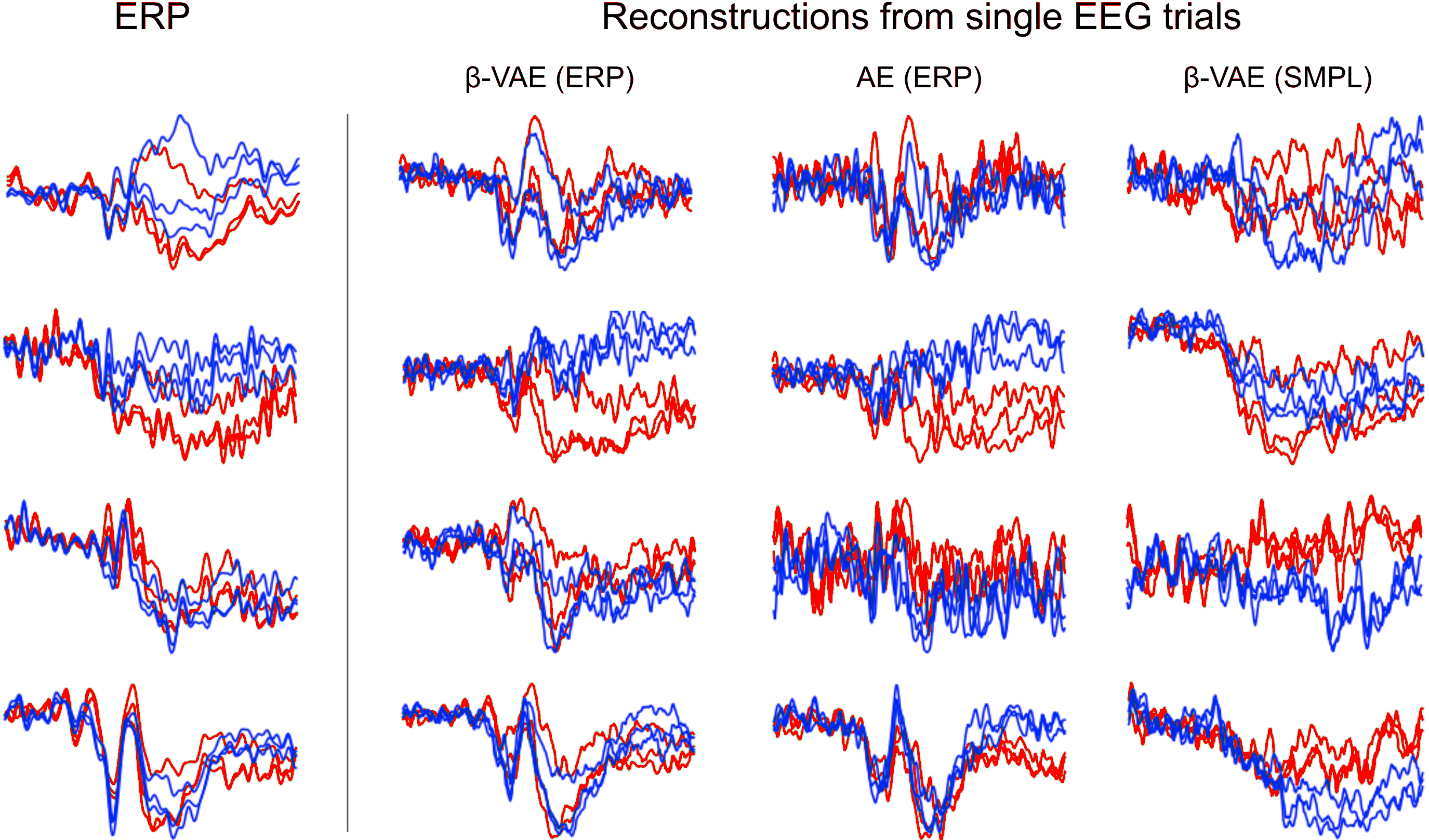}
 \caption{Reconstructions from single EEG trials using two disentangled $\beta$-VAEs pre-trained on either ERPs or single EEG sample trajectories (SMPL), as well as an AE pre-trained on ERPs (AE ERP). Models pre-trained on ERPs are able to reconstruct ERP-like trajectories even from single EEG samples, as demonstrated by the closer similarity of the $\beta$-VAEs (ERP) and AE (ERP) reconstructions to the ground truth ERPs (leftmost column). The ground truth ERPs were obtained by averaging on average 37 single EEG trials, including the single EEG trajectories that were used as inputs to the models.}
 \label{fig_sample_recs_3}
\end{figure}

\begin{figure}[h!]
 \centering
 \includegraphics[width=1.\textwidth]{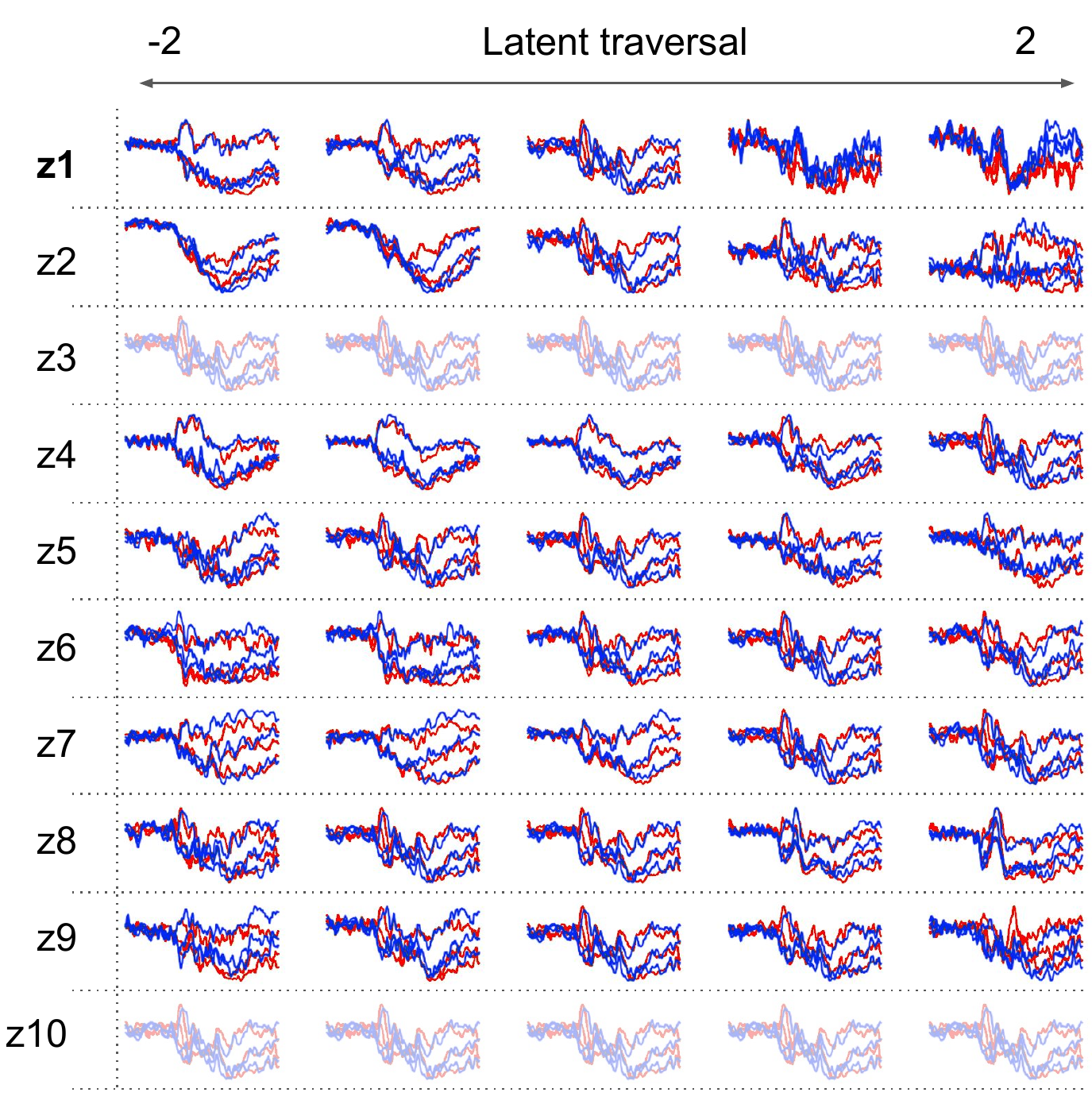}
 \caption{Latent traversals of a well disentangled pre-trained $\beta$-VAE with 8 ``informative'' dimensions (UDR=0.36) indicated by an arrow in Fig.~\ref{fig_udr_scores}. Each row reconstructs an ERP trajectory as the value of each latent dimension is traversed between [-2, 2] while keeping the values of all other latents fixed. Greyed out latents are ``uninformative'' -- the model effectively switched them off by setting their inferred posterior to the unit Gaussian prior. Bold latent was identified by SCAN to be associated with depression.}
 \label{fig_traversals_sup_1}
\end{figure}

\begin{figure}[h!]
 \centering
 \includegraphics[width=1.\textwidth]{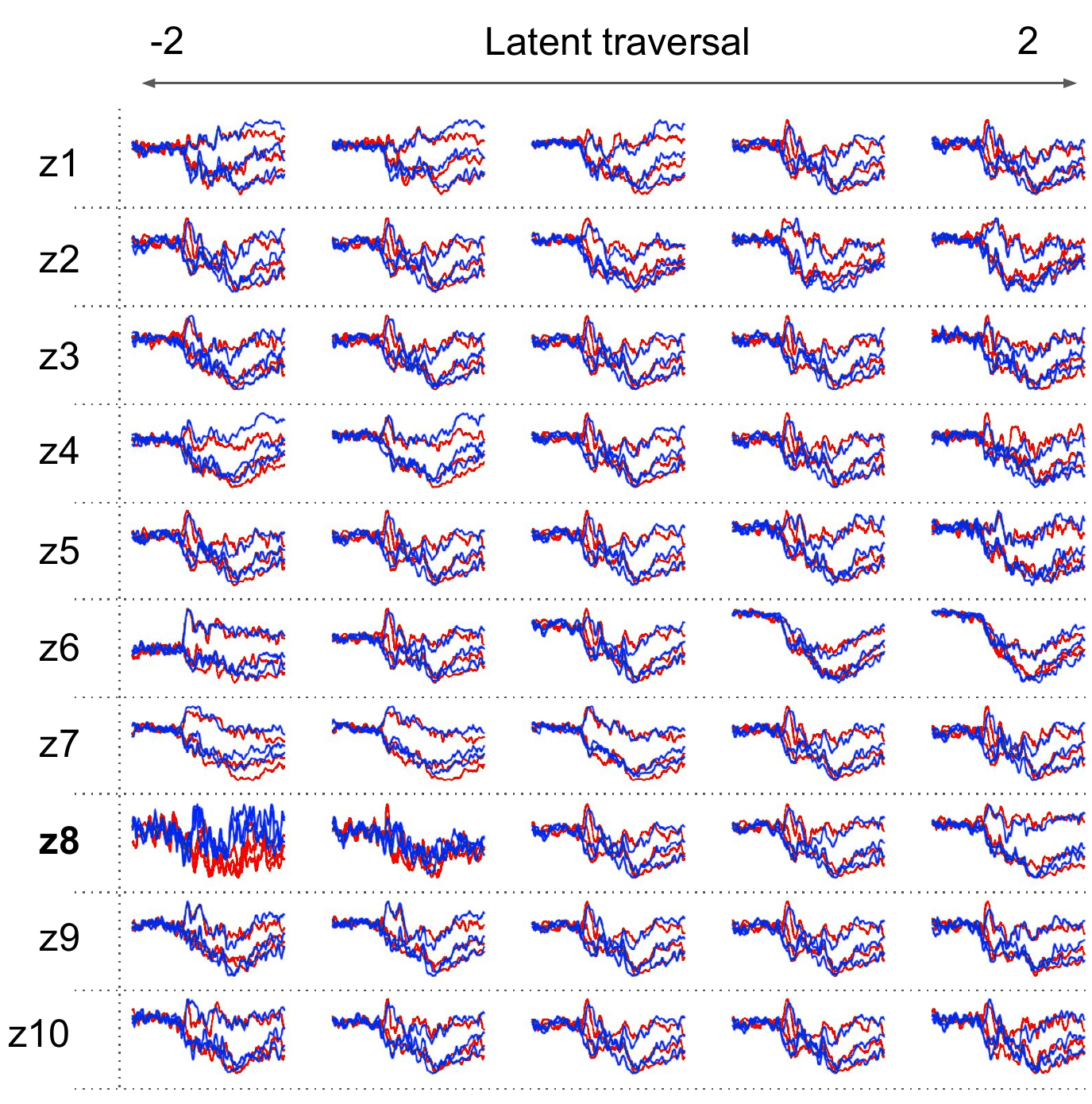}
 \caption{Latent traversals of a well disentangled pre-trained $\beta$-VAE with 9 ``informative'' dimensions (UDR=0.36) indicated by an arrow in Fig.~\ref{fig_udr_scores}. Each row reconstructs an ERP trajectory as the value of each latent dimension is traversed between [-2, 2] while keeping the values of all other latents fixed. Greyed out latents are ``uninformative'' -- the model effectively switched them off by setting their inferred posterior to the unit Gaussian prior. Bold latent was identified by SCAN to be associated with depression.}
 \label{fig_traversals_sup_2}
\end{figure}

\begin{figure}[h!]
 \centering
 \includegraphics[width=1.\textwidth]{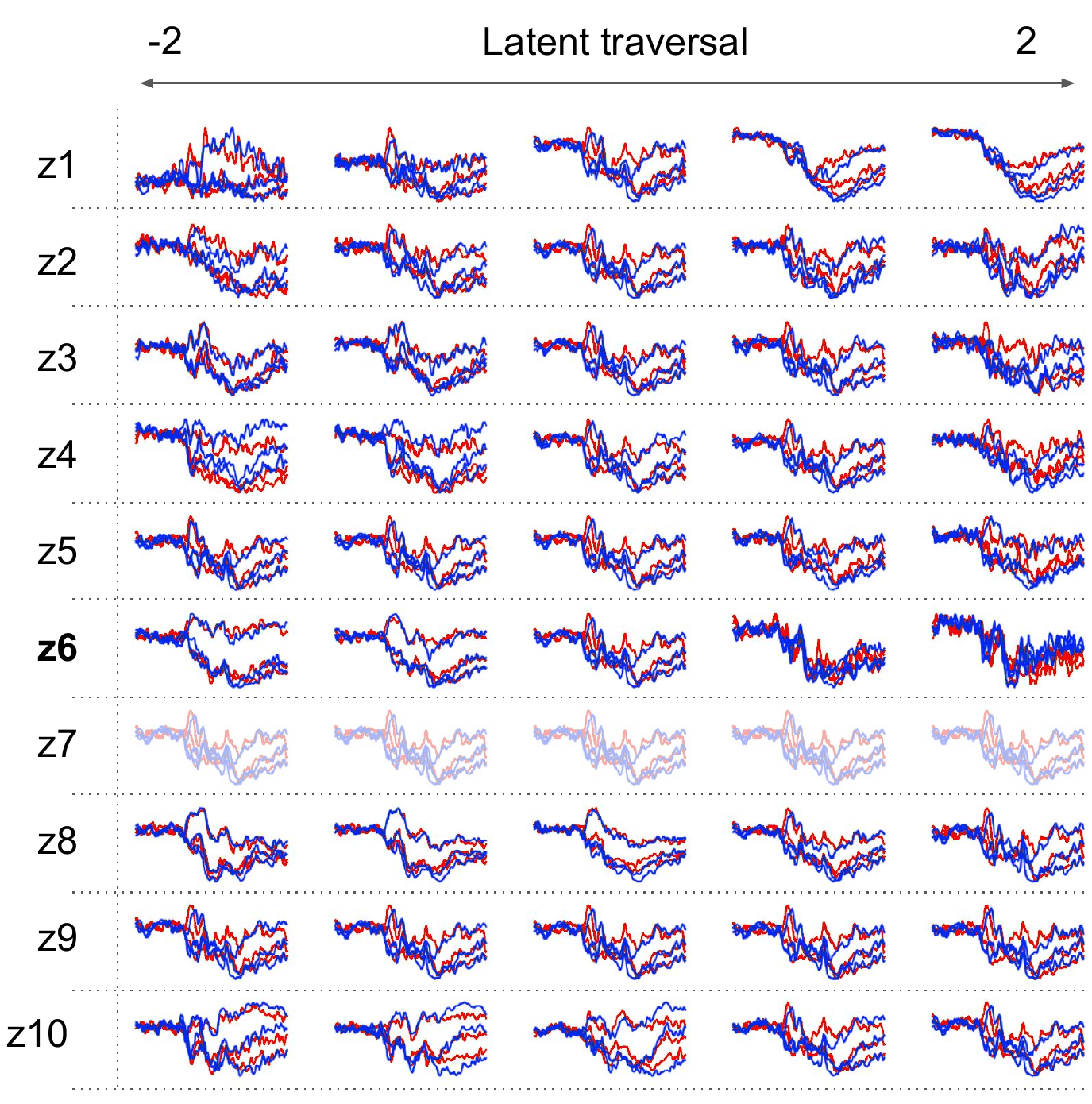}
 \caption{Latent traversals of a well disentangled pre-trained $\beta$-VAE with 10 ``informative'' dimensions (UDR=0.31) indicated by an arrow in Fig.~\ref{fig_udr_scores}. Each row reconstructs an ERP trajectory as the value of each latent dimension is traversed between [-2, 2] while keeping the values of all other latents fixed. Bold latent was identified by SCAN to be associated with depression.}
 \label{fig_traversals_sup_3}
\end{figure}

\end{document}